\documentclass[journal,twocolumn,10pt]{IEEEtran} 
\usepackage{lipsum} 
\usepackage[tight,footnotesize]{subfigure}
\usepackage{enumitem} 
\usepackage{float}
\usepackage{multicol,multienum}
\usepackage{breqn}
\usepackage{stfloats}
\usepackage{multirow}
\usepackage{diagbox}
\usepackage{hyperref}
\usepackage{bm}
\usepackage{cite}
\usepackage{graphicx}
\usepackage{tabularx}
\usepackage[ruled,vlined]{algorithm2e} 
\usepackage{booktabs}

\usepackage[tight,footnotesize]{subfigure}
\usepackage{url}
\usepackage{doi}
\usepackage{amssymb}

\usepackage[switch]{lineno}
\usepackage[todonotes={textsize=scriptsize}]{changes}
\makeatletter
\@namedef{Changes@AuthorColor}{red}
\colorlet{Changes@Color}{red}
\makeatother
\ifCLASSINFOpdf
\else
\fi
\begin{document}
%
\title{Privacy-preserving Traffic Flow Prediction: A Federated Learning Approach }
%
%
%

\author{Yi~Liu,~\IEEEmembership{Student Member,~IEEE,}
        James~J.Q. Yu,~\IEEEmembership{Member,~IEEE,}
        Jiawen~Kang, 
        Dusit~Niyato,~\IEEEmembership{Fellow,~IEEE,}
        Shuyu~Zhang
\thanks{Yi Liu, James J.Q. Yu, and Shuyu Zhang are with the Department of Computer Science and Engineering, Southern University of Science and Technology, Shenzhen, China. Yi Liu is also with the School of Data Science and Technology, Heilongjiang University, Harbin, China (email: 97liuyi@ieee.org; yujq3@sustech.edu.cn; 11712122@mail.sustech.edu.cn).}
\thanks{Jiawen Kang and Dusit Niyato are with the School of Computer Science and Engineering, Nanyang Technological University, Singapore (e-mail: kavinkang@ntu.edu.sg; dniyato@ntu.edu.sg).}
\thanks{This work is supported by the General Program of Guangdong Basic and Applied Basic Research Foundation under grant No. 2019A1515011032.}
}

%
%

\markboth{IEEE Internet of Things Journal}%
{}
%



\maketitle

\begin{abstract}
Existing traffic flow forecasting approaches by deep learning models achieve excellent success based on a large volume of datasets gathered by governments and organizations. However, these datasets may contain lots of user's private data, which is challenging the current prediction approaches as user privacy is calling for the public concern in recent years. Therefore, how to develop accurate traffic prediction while preserving privacy is a significant problem to be solved, and there is a trade-off between these two objectives. To address this challenge, we introduce a privacy-preserving machine learning technique named federated learning and propose a Federated Learning-based Gated Recurrent Unit neural network algorithm (FedGRU) for traffic flow prediction. FedGRU differs from current centralized learning methods and updates universal learning models through a secure parameter aggregation mechanism rather than directly sharing raw data among organizations. In the secure parameter aggregation mechanism, we adopt a Federated Averaging algorithm to reduce the communication overhead during the model parameter transmission process. Furthermore, we design a Joint Announcement Protocol to improve the scalability of FedGRU. We also propose an ensemble clustering-based scheme for traffic flow prediction by grouping the organizations into clusters before applying FedGRU algorithm. Through extensive case studies on a real-world dataset, it is shown that FedGRU's prediction accuracy is 90.96\% higher than the advanced deep learning models, which confirm that FedGRU can achieve accurate and timely traffic prediction without compromising the privacy and security of raw data.

\end{abstract}


%
\IEEEpeerreviewmaketitle

\section{Introduction}
%
%
%
%
\IEEEPARstart{C}{ontemporary} urban residents, taxi drivers, business sectors, and government agencies have a strong need of accurate and timely traffic flow information \cite{ref-1} as these road users can utilize such information to alleviate traffic congestion, control traffic light appropriately, and improve the efficiency of traffic operations \cite{ref-2,ref-66,ref-67}. Traffic flow information can also be used by people to develop better travelling plans. Traffic Flow Prediction (TFP) is to provide such traffic flow information by using historical traffic flow data to predict future traffic flow. TFP is regarded as a critical element for the successful deployment of Intelligent Transportation System (ITS) subsystems, particularly the advanced traveler information, online car-hailing, and traffic management systems.

In TFP, centralized machine learning methods are typically utilized to predict traffic flow by training with sufficient sensor data, e.g., from mobile phones, cameras, radars, etc. For example, Convolutional Neural Networks (CNN), Recurrent Neural Networks (RNN) and their variants have achieved gratifying results in predicting traffic flow in the literature. Such learning methods typically collaboratively require sharing data among public agencies and private companies. Indeed, in recent years, the general public witnessed partnerships among public agencies and mobile service providers such as DiDi Chuxing, Uber, and Hellobike. These partnerships extend the capability and services of companies that provide real-time traffic flow forecasting, traffic management, car sharing, and personal travel applications \cite{ref-64}. 

\begin{figure}[!t]
	\centering
	\includegraphics[width=1\linewidth]{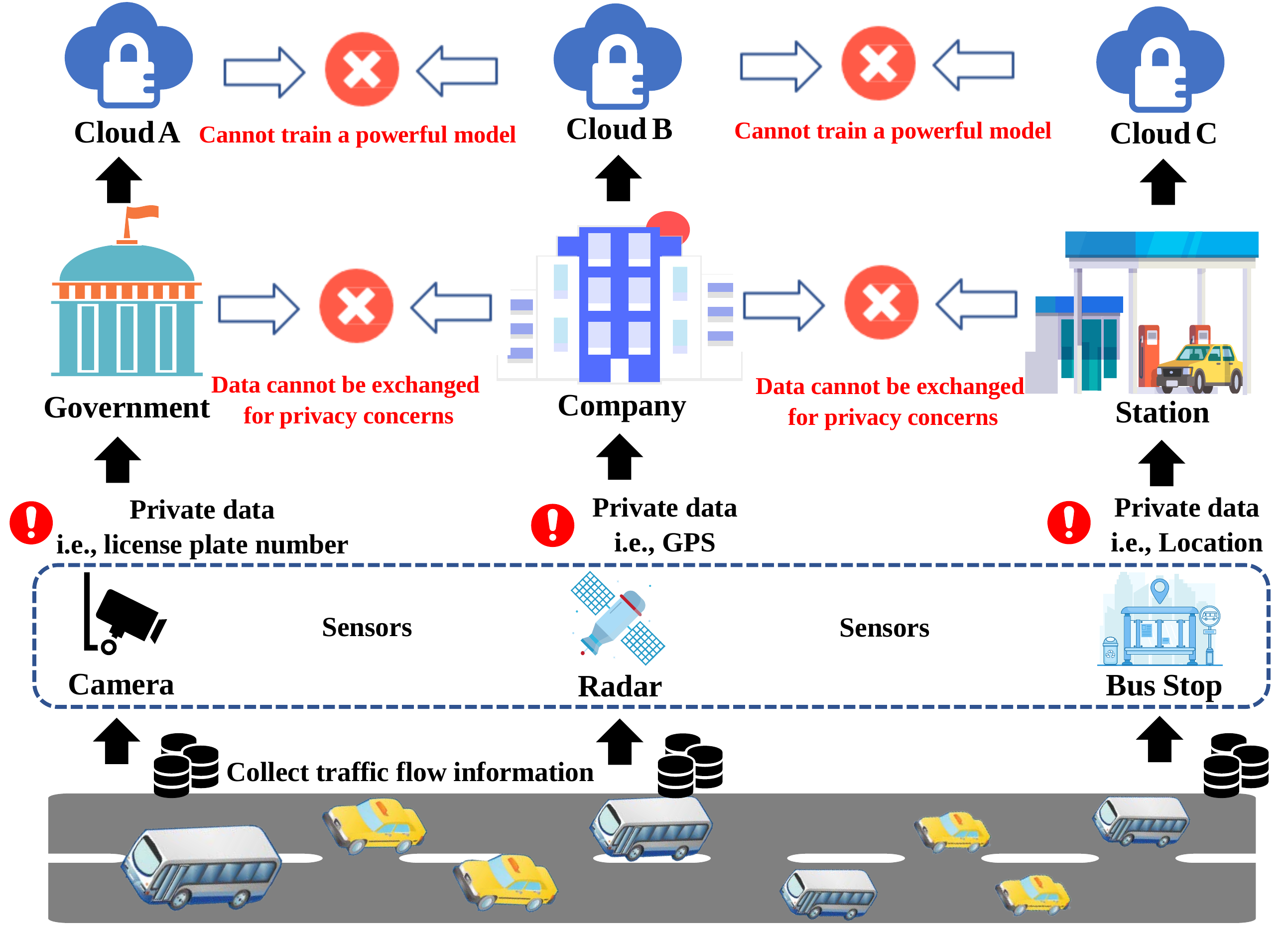}
	\caption{Privacy and security problems in traffic flow prediction.}
	\label{fig-1}
\end{figure}

Nonetheless, it is often overlooked that the data may contain sensitive private information, which leads to potential privacy leakage. As shown in Fig. \ref{fig-1}, there are some privacy issues in the traffic flow prediction context. For example, road surveillance cameras capture vehicle license plate information when monitoring traffic flow, which may leak user private information \cite{ref-61}  When different organizations use data collected by sensors to predict traffic flow, the collected data is stored in different clouds and should not be exchanged for privacy preservation. These make it challenging to train an effective model with this valuable data. While the assumption is widely made in the literature, the acquisition of massive user data is not possible in real applications respecting privacy. Furthermore, Tesla Motors leaked the vehicle's location information when using the vehicle's GPS data to achieve traffic prediction, which would cause many security risks to the owner of the vehicle\footnote{\href{https://www.anquanke.com/post/id/197750}{https://www.anquanke.com/post/id/197750}}. In EU, Cooperative ITS (C-ITS)\footnote{\href{https://new.qq.com/omn/20180411/20180411A1W9FI.html}{https://new.qq.com/omn/20180411/20180411A1W9FI.html}} service providers must provide clear terms to end-users in a concise and accessible form so that users can agree to the processing of their personal data \cite{ref-63}. Therefore, it is important to protect privacy while predicting traffic flow.
To predict traffic flow in ITS without compromising privacy, reference \cite{ref-7} introduced a privacy control mechanism based on ``$k$-anonymous diffusion,'' which can complete taxi order scheduling without leaking user privacy. Le Ny \textit{et al.} proposed a differentially private real-time traffic state estimator system to predict traffic flow in \cite{ref-43}. However, these privacy-preserving methods cannot achieve the trade-off between accuracy and privacy, rendering subpar performance. Therefore, we need to seek an effective method to accurately predict traffic flow under the constraint of privacy protection.


To address the data privacy leakage issue, we incorporate a privacy-preserving machine learning technique named Federated Learning (FL) \cite{ref-16} for TFP in this work. In FL, distributed organizations cooperatively train a globally shared model through their local data without exchanging the raw data. To accurately predict traffic flow, we propose an enhanced federated learning algorithm with a Gated Recurrent Unit neural network (FedGRU) in this paper, where GRU is an advanced time series prediction model that can be used to predict traffic flow.  Through FL and its aggregation mechanism \cite{ref-11}, FedGRU aggregates model parameters from different geographically located organizations to build a global deep learning model under privacy well-preserved conditions. Furthermore, contributed by the outstanding data regression capability of GRU neural networks, FedGRU can achieve accurate and timely traffic flow prediction for different organizations. 
The major contributions of this paper are summarized as follows:
\begin{itemize}
	
\item Unlike existing algorithms, we propose a novel privacy-preserving algorithm that integrates emerging federated learning with a practical GRU neural network for traffic flow prediction. Such an algorithm provides reliable data privacy preservation through a locally training model without raw data exchange.
\item To improve the scalability and scalabiligy of federated learning in traffic flow prediction, we design an improved Federated Averaging (FedAVG) algorithm with a Joint-Announcement protocol in the aggregation mechanism. This protocol uses random sub-sampling for participating organizations to reduce the communication overhead of the algorithm, which is particularly suitable for large-scale and distribution prediction.
\item Based on FedGRU algorithm, we develop an ensemble clustering-based FedGRU scheme to integrate the optimal global model and capture the spatio-temporal correlation of traffic flow data, thereby further improving the prediction accuracy.
\item We conduct extensive experiments on a real-world dataset to demonstrate the performance of the proposed schemes for traffic flow prediction compared to non-federated learning methods.

\end{itemize}
	
The remainder of this paper is organized as follows. Section \ref{sec-5} reviews the literature on short-term TFP and privacy research in ITS. Section \ref{sec-2} defines the Centralized TFP Learning problem and Federated TFP Learning problem, and proposes a security parameter aggregation mechanism. Section \ref{sec-3} presents FedGRU algorithm and ensemble clustering-based FedGRU algorithm. In FedGRU, we introduce FedAVG algorithm, Joint-Announcement Protocol in detail. Section \ref{sec-4} and Section \ref{diss} discusse the experimental results. Concluding remarks are described in Section \ref{sec-6}.

\section{Related Work}\label{sec-5}
\subsection{Traffic Flow Prediction}
Traffic flow forecasting has always been a hot issue in ITS, which serves as functions of real-time traffic control and urban planning. Although researchers have proposed many new methods, they can generally be divided into two categories: parametric models and non-parametric models.
\subsubsection{Parametric models} Parametric models predict future data by capturing existing data feature within its parameters. M. S. Ahmed\textit{ et al.} in \cite{ref-24} proposed the Autoregressive Integrated Moving Average (ARIMA) model in the 1970s to predict short-term freeway traffic. Since then, many researchers have proposed variants of ARIMA such as Kohonen-ARIMA (KARIMA) \cite{ref-25}, subset ARIMA \cite{ref-26}, seasonal ARIMA \cite{ref-27}, etc. These models further improve the accuracy of TFP by focusing on the statistical correlation of the data. Parametric models have several advantages. First of all, such parametric models are highly transparent and interpreted for easy human understanding. Second, these solutions usually take less time than non-parametric models. However, these solutions suffer from low model express ability, rarely solutions to achieve accurate and timely TFP.

\subsubsection{Non-parametric models} With the improvement of data storage and computing, non-parametric models have achieved great success in TFP \cite{ref-28}. Davis and Nihan \textit{et al.} in \cite{ref-29} proposed $k$-NN model for short-term traffic flow prediction. Lv \textit{et al.} in \cite{ref-1} first applied the stacked autoencoder (SAE) model to TFP. Furthermore, SAE adopts a hierarchical greedy network structure to learn non-linear features and has better performance than Support Vector Machines (SVM)\cite{ref-30} and Feed-forward Neural Network (FFNN) \cite{ref-31}. Considering the timing of the data, Ma \textit{et al. }in \cite{ref-32} and Tian \textit{et al.} in \cite{ref-33} applied Long Short-Term Memory (LSTM) to achieve accurate and timely TFP. Fu \textit{et al.}in \cite{ref-15} first proposed GRU neural network methods for TFP. In recent years, due to the success of convolutional networks and graph networks, Yu\textit{ et al. }in \cite{ref-35,ref-34} proposed graph convolutional generative autoencoder to address the real-time traffic speed estimation problem. 

\subsection{Privacy Issues for Intelligent Transportation Systems }
In ITS, many models and methods rely on training data from users or organizations. However, with the increasing privacy awareness, direct data exchange among users and organizations is not permitted by law. Matchi \textit{et al.} in \cite{ref-36} developed privacy-preserving service to compute meeting points in ride-sharing based on secure multi-party computing, so that each user remains in control of his location data. Brian \textit{et al.} \cite{ref-37} designed a data sharing algorithm based on information-theoretic $k$-anonymity. This data sharing algorithm implements secure data sharing by $k$-anonymity encryption of the data. The authors in \cite{ref-55} proposed a privacy-preserving transportation traffic measurement scheme for cyber-physical road systems by using maximum-likelihood estimation (MLE) to obtain the prediction result. Reference \cite{ref-56} presents a system based on virtual trip lines and an associated cloaking technique to achieve privacy-protected traffic flow monitoring. To avoid leaking the location of vehicles while monitoring traffic flow, \cite{ref-57} proposed a privacy-preserving data gathering scheme by using encrypt methods. Nevertheless, these approaches have two problems: 1) they respects privacy at the expense of accuracy; 2) they cannot properly handle a large amount of data within limit time \cite{ref-10}. Besides, the EU has promulgated General Data Protection Regulation (GDPR), which means that as long as the organization has the possibility of revealing privacy in the data sharing process, the data transaction violates the law \cite{ref-65}. Therefore, we need to develop new methods to adapt to the general public with a growing sense of privacy.


In recent years, Federated Learning (FL) models have been used to analyze private data because of its privacy-preserving features. FL is to build machine-learning models based on datasets that are distributed across multiple devices while preventing data leakage \cite{ref-42}. Bonawitz \textit{et al.} in\cite{ref-38} first applied FL to decentralized learning of mobile phone devices without compromising privacy. To ensure the confidentiality of the user's local gradient during the federated learning process, the author in \cite{ref-58} proposed VerifyNet, which is the first framework to protect privacy and verifiable federated learning. Reference \cite{ref-59} used a multi-weight subjective logic model to design a reputation-based device selection scheme for reliable federated learning. The authors in \cite{ref-60} applied the federated learning framework to edge computing to achieve privacy-protected data analysis. Nishio \textit{et al.} in\cite{ref-39} applied FL to mobile edge computing and proposed the FedCS protocol to reduce the time of the training process. Chen \textit{et al.} in \cite{ref-40} combined transfer learning \cite{ref-41} with FL to propose FedHealth to be applied to healthcare. Yang \textit{et al.} in \cite{ref-42} introduced Federated Machine Learning, which can be applied to multiple applications in smart cities such as energy demand forecasting.

Although researchers have proposed some privacy-preserving methods to predict traffic flow, they do not comply with the requirements of GDPR. In this paper, we explore a privacy-preserving method  FL with GRU for traffic flow prediction. 

\section{PROBLEM DEFINITION}\label{sec-2}
We use the term ``organization" throughout the paper to describe entities in TFP, such as urban agencies, private companies, and detector stations. We use the term ``client'' to describe computing nodes that correspond to one or multiple sensors in FL and use the term ``device'' to describe the sensor in the organizations. Let $\mathcal{C}  = \{ {C_1},{C_2},\cdots,C_n\}$ and $\mathcal{O}  = \{ {O_1},{O_2},\cdots,{O_m}\}$ denote the client set and organization set in ITS, respectively. In the context of traffic flow prediction, we treat organizations as clients in the definition of federated learning. This equivalency does not undermine the privacy preservation constraint of the problem and the federated learning framework. Each organization has $k_i$ devices and their respective database $D_i$. We aim to predict the number of vehicles with historical traffic flow information from different organizations without sharing raw data and privacy leakage. We design a secure parameter aggregation mechanism as follows:

{\textbf{Secure Parameter Aggregation Mechanism:}} \textit{Detector station $O_i$ has $N$ devices, and the traffic flow data collected by the $N$ devices constitute a database $D_i$. The deep learning model constructed in $O_i$ calculates updated model parameters $ {p_i} $ using the local training data from $D_i$. When all detector stations finish the same operation, they upload their respective $p_i$ to the cloud and aggregate a new global model.}\label{defi-1}

\begin{figure}[!t]
	\centering
	\includegraphics[width=1\linewidth]{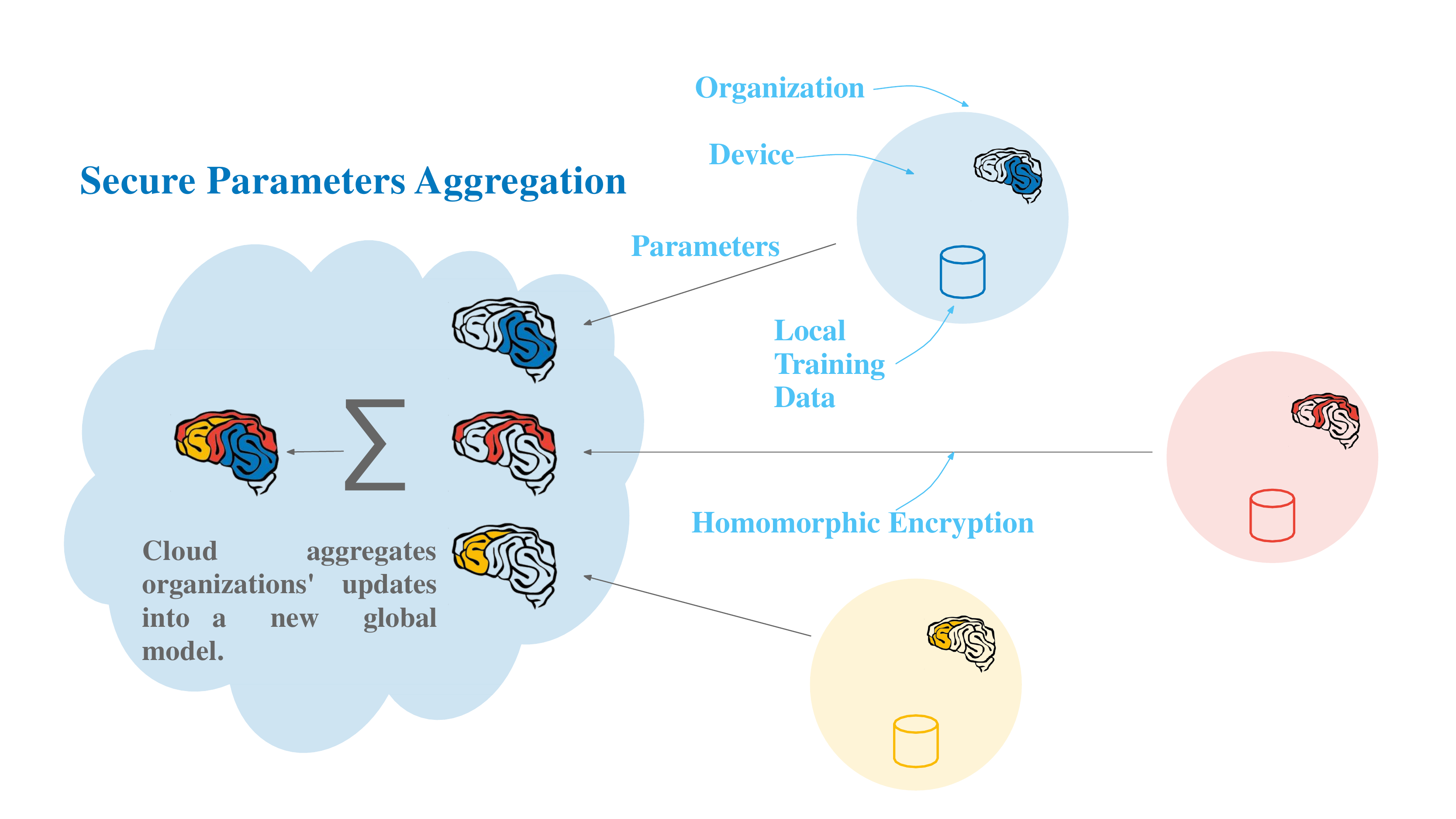}
	\caption{Secure parameter aggregation mechanism. }
	\label{fig-3}
\end{figure}
According to Secure Parameters Aggregation, no traffic flow data is exchanged among different detector stations. The cloud aggregates organizations' submitted parameters into a new global model without exchanging data. (As shown in Fig. \ref{fig-3}) 

In this paper, $t$ and $v_t$ represent the $t$-th timestamp in the time-series and traffic flow at the $t$-th timestamp, respectively.  Let $f( \cdot )$ be the traffic flow prediction function, the definitions of privacy, centralized, and federated TFP learning problems as follows:

\textbf{Information-based Privacy:} \textit{Information-based privacy defines privacy as preventing direct access to private data. This data is associated with the user's personal information or location. For example, a mobile device that records location data allows weather applications to directly access the location of the current smartphone, which actually violates the information-based privacy definition \cite{ref-11}. In this work, every device trains its local model by using local dataset instead of sharing the dataset and upload the updated gradients (i.e., parameters) to the cloud.}

\textbf{Centralized TFP Learning:} \textit{Given organizations $\mathcal{ O}$, each organization's devices $k_i$, and an aggregated database $D = {D_1} \cup {D_2} \cup {D_3} \cup \cdots\cup {D_N}$, the centralized TFP problem is to calculate ${v_{t + s}} = f(t + s,D)$, where $s$ is the prediction window after $t$. }

\textbf{Federated TFP Learning:} \textit{Given organizations $\mathcal{O}$ and each organization's devices $k_i$, and their respective database $D_i$, the federated TFP problem is to calculate ${v_{t + s}} = {f_i}(t + s,{D_i})$ where ${f_i}( \cdot , \cdot )$ is the local version of $f( \cdot , \cdot )$ and $s$ is the prediction window after $t$. Subsequently, the produced results are aggregated by a secure parameter aggregation mechanism.}

\section{METHODOLOGY}\label{sec-3}
Traditional centralized learning methods consist of three steps: data processing, data fusion, and model building. In the traditional centralized learning context, data processing means that data feature and data label need to be extracted from the original data (e.g. text, images, and application data) before performing the data fusion operation. Specifically, data processing includes sample sampling, outlier removal, feature normalization processing, and feature combination. For the data fusion step,  traditional learning models  directly share data among all parties to obtain a global database for training. Such a centralized learning approach faces the challenge of new data privacy laws and regulations as organizations may disclose privacy and violate laws such as GDPR when sharing data. FL is introduced into this context to address the above challenges. However, existing FL frameworks typically employ simple machine learning models such as XGBoost and decision trees rather than complicated deep learning models \cite{ref-16,ref-18}. Because such models need to upload a large number of parameters to the cloud in FL framework, it leads to expensive communication overhead which can cause training failures for a single model or a global model \cite{ref-12,ref-8}. Therefore, FL framework needs to develop a new parameter aggregation mechanism for deep learning models to reduce communication overhead.

In this section, we present two approaches to predict traffic flow, including FedGRU and clustering-based FedGRU algorithms. Specifically, we describe an improved Federated Averaging (FedAVG) algorithm with a Joint-Announcement protocol in the aggregation mechanism to reduce the communication overhead. This approach is useful to implement in the following particular scenarios.

\subsection{Federated Learning and Gated Recurrent Unit}\label{gru}
Federated Learning (FL) \cite{ref-16} is a distributed machine learning (ML) paradigm that has been designed to train ML models without compromising privacy. With this scheme, different organizations can contribute to the overall model training while keeping the training data locally. 

Particularly, FL problem involves learning a \textit{single} and \textit{globally} predicted model from the database separately stored in dozens of or even hundreds of organizations \cite{ref-17,ref-50}. We assume that a set $\mathcal{K}$ of $K$ device stores its local dataset $\mathcal{D}_k$ of size $D_k$. So we can define the local training dataset size $D = \sum\nolimits_{k = 1}^K {{D_k}} $. In a typical deep learning setting, given a set of input-output pairs $\{ {x_i},{y_i}\} _{i = 1}^{{D_k}}$, where the input sample vector with $d$ features is ${x_i} \in {\mathbb{R}^d}$, and the labeled output value for the input sample $x_i$ is ${y_i} \in \mathbb{R}$. If we input the training sample vector $x_i$ (e.g., the traffic flow data), we need to find the model parameter vector $\omega  \in {\mathbb{R}^d}$ that characterrizes the output $y_i$ (e.g., the value output of the traffic flow data) with loss function ${f_i}(\omega )$ (e.g., ${f_i}(\omega ) = \frac{1}{2}(x_i^T\omega  - {y_i})$). Our goal is to learn this model under the constraints of local data storage and processing by devices in the organization with a secure parameter aggregation mechanism. The loss function on the data set of device $k$ is defined as:

\begin{equation}\label{eq-local}
		{J_k}(\omega ): = \frac{1}{{{D_k}}}\sum\nolimits_{i \in {D_k}} {{f_i}(\omega ) \lambda h(\omega )} ,
\end{equation}
where $\omega  \in {{\mathbb {R}}^d}$  is the local model parameter, $\forall \lambda  \in [0,1]$ , and $h( \cdot )$ is a regularizer function. ${D_k}$ is used in Eq. (\ref{eq-local}) to illustrate that the local model in device $k$ needs to learn every sample in the local data set.

At the cloud, the global predicted model problem can be represented as follows:
\begin{equation}\label{eq-local-2}
		\arg \mathop {\min }\limits_{\omega  \in {{\mathrm R}^d}} J(\omega ),J(\omega ) = \sum\nolimits_{k = 1}^K {\frac{{{H_k}}}{D}} {F_k}(\omega ),
\end{equation}
we recast the global predicted model problem in (\ref{eq-4}) as follows:
\begin{equation}\label{eq-6}
		\arg \mathop {\min }\limits_{\omega  \in {{\mathrm R}^d}} J(\omega ): = \sum\nolimits_{k = 1}^K {\frac{{\sum\nolimits_{i \in {D_k}} {{f_i}(\omega ) + \lambda h(\omega )} }}{D}}.
\end{equation}
The Eq. (\ref{eq-local-2})-(\ref{eq-6}) illustrates the global model aggregation of model update aggregations uploaded by each device to obtain updates.

For the TFP problem, we regard GRU neural network model as the local model in Eq. (\ref{eq-local}). Cho \textit{et al.} in \cite{ref-14} proposed the GRU neural network in 2014, which is a variant of RNN that handles time-series data. GRU is different from RNN is that it adds a ``Processor'' to the algorithm to judge whether the information is useful or not. The structure of the processor is called ``Cell.'' A typical structure of GRU cell uses two data ``gates'' to control the data from processor: reset gate $r$ and update gate $z$. 

Let $X = \{ {x_1},{x_2},\cdots,{x_n}\} $, $Y = \{ {y_1},{y_2},\cdots,{y_n}\} $, and $H = \{ {h_1},{h_2},\cdots,{h_n}\} $ be the input time series, output time series and the hidden state of the cells, respectively. At time step $t$, the value of update gate $z_t$ is expressed as:
\begin{equation}\label{eq-2}
{z_t} = \sigma ({W^{(z)}}{x_t} + {U^{(z)}}{h_{t - 1}}),
\end{equation}
where $x_t$ is the input vector of the $t$-th time step, ${W^{(z)}}$ is the weight matrix, and $h_{t-1}$ holds the cell state of the previous time step $t-1$. The update gate aggregates ${W^{(z)}}{x_t}$ and ${U^{(z)}}{h_{t - 1}}$, then maps the results in $(0,1)$ through a Sigmoid activation function. Sigmoid activation can transform data into gate signals and transfer them to the hidden layer. The reset gate $r_t$ is computed similarly to the update gate:
\begin{equation}\label{eq-3}
{r_t} = \sigma ({W^{(z)}}{x_t} + {U^{(r)}}{h_{t - 1}}).
\end{equation}

The candidate activation ${h_t}'$ is denoted as:
\begin{equation}\label{eq-4}
{h_t}' = \tanh (W{x_t} + {r_t} \odot U{h_{t - 1}}),
\end{equation}
where ${r_t} \odot U{h_{t - 1}}$ represents the Hadamard product of $r_t$ and $U{h_{t - 1}}$. The $\tanh $ activation function can map the data to the range of (-1,1), which can reduce the amount of calculations and prevent gradient explosions.

The final memory of the current time step $t$ is calculated as follows:
\begin{equation}\label{eq-5}
{h_t} = {z_t} \odot {h_{t - 1}} + (1 - {z_t}) \odot {h_t}'.
\end{equation}

\subsection{Privacy-preserving Traffic Flow Prediction Algorithm}
\begin{figure}[!t]
	\centering
	\includegraphics[width=1\linewidth]{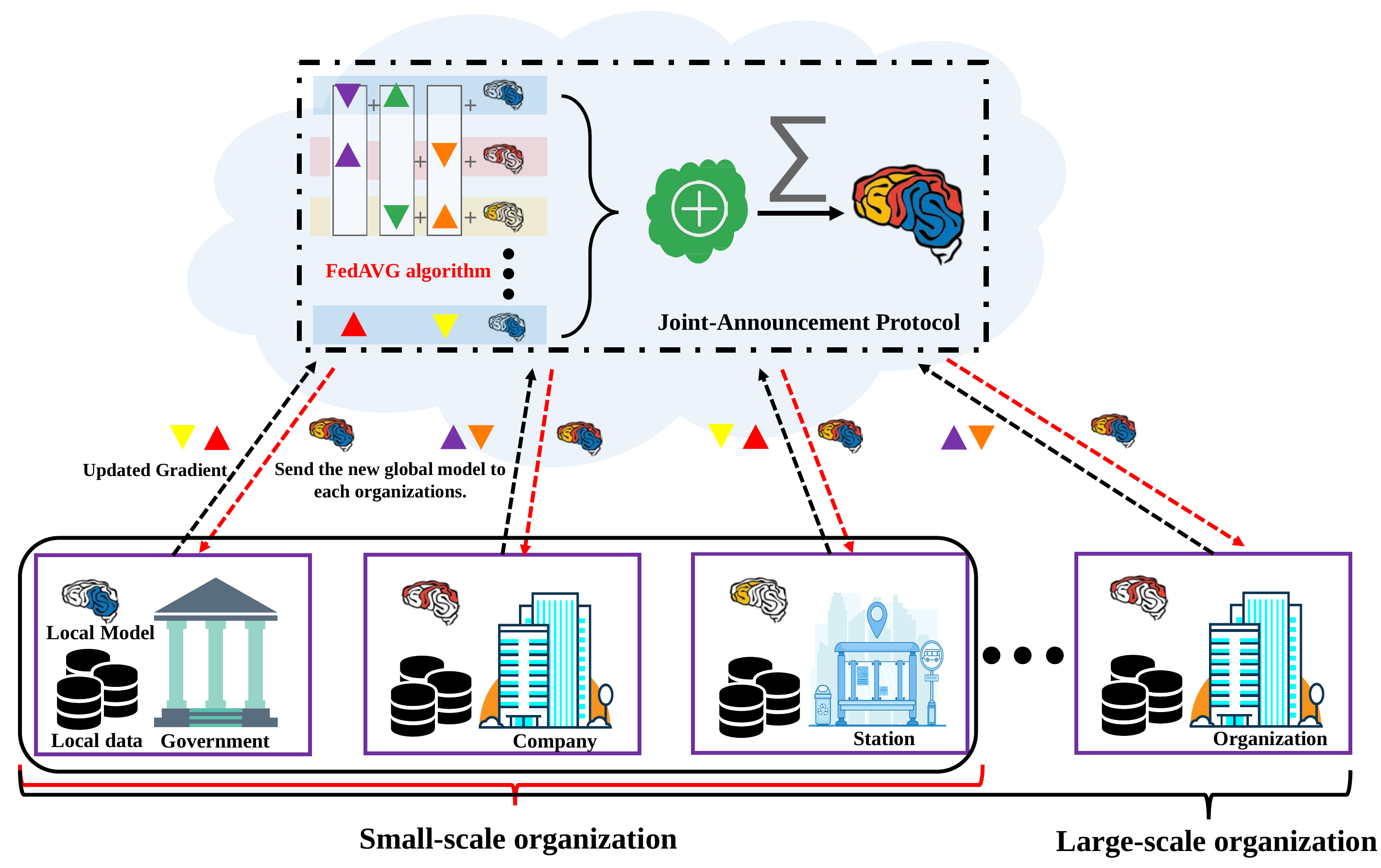}
	\caption{Federated learning-based traffic flow prediction architecture. Note that when the organization in the architecture is small-scale, we will use the FedAVG algorithm to calculate the update, and when the organization is large-scale, we will use the Joint-Announcement Protocol to calculate the update by subsampling the organizations. Details will be described in detail in following sub-subsection \ref{pro}.  }
	\label{fig-6}
\end{figure}

Since centralized learning methods use central database $\mathcal{D}$ to merge data from organizations and upload the data to the cloud, it may lead to expensive communication overhead and data privacy concerns. To address these issues, we propose a privacy-preserving traffic flow prediction algorithm FedGRU as shown in Fig. \ref{fig-6}. Firstly, we introduce FedAVG algorithm as the core of the secure parameter aggregation mechanism to collect gradient information from different organizations. Secondly, we design an improved FedAVG algorithm with a Joint-Announcement protocol in the aggregation mechanism. This protocol uses random sub-sampling for participating organizations to reduce the communication overhead of the algorithm, which is particularly suitable for larger-scale and distribution prediction. Finally, we give the details of FedGRU, which is a prediction algorithm combining FedAVG and Joint-Announcement protocol.
\subsubsection{FedAVG algorithm}\label{avg}
A recognized problem in federated learning is the limited network bandwidth that bottlenecks cloud-aggregated local updates from the organizations. To reduce the communication overhead,  each client uses its local data to perform gradient descent optimization on the current model. Then the central cloud performs a weighted average aggregation of the model updates uploaded by the clients.  As shown in Algorithm \ref{al-2}, FedAVG consists of three steps:
\begin{enumerate}[label=(\roman*)]
	\item The cloud selects volunteers from organizations $\mathcal{ O}$ to participate in this round of training and broadcasts global model $\omega ^o$ to the selected organizations;
	\item Each organization $o$ trains data locally and updates ${\omega _{t}^o}$ for $E$ epochs of SGD with mini-batch size $B$ to obtain $\omega_{t+1}^o$, i.e., $\omega _{t + 1}^o \leftarrow$ $\mathrm{LocalUpdate}$$(o,\omega _{t}^o)$;
	\item The cloud aggregates each organization's $\omega_{t+1}$ through a secure parameter aggregation mechanism.
\end{enumerate}
\begin{algorithm}[t]\label{al-1}
	\caption{Federated Averaging (FedAVG) Algorithm.}
	\LinesNumbered 
	\KwIn{Organizations $\mathcal{O} = \{ {O_1},{O_2},\cdots,{O_N}\} $. $B$ is the local mini-batch size, $E$ is the number of local epochs, $\alpha$ is the learning rate, $\nabla \mathcal{L}(\cdot ;\cdot)$ is the gradient optimization function.}
	\KwOut{Parameter $\omega$.}
	Initialize $ \omega^{0}$ (Pre-trained by a public dataset)\;
	\ForEach{round $t = 1,2,\cdots$}{
	$\{ {O_v}\}  \leftarrow $ select volunteer from organizations $\mathcal{O}$ participate in this round of training\;
	Broadcast global model $\omega ^o$ to organization in $\{ {O_v}\}$\;
	 \ForEach{organization $o \in \{O_v\}$ \textbf{in parallel}}
	 {
	 Initialize $\omega _t^o = {\omega ^o}$\;
	 	 $\omega_{t+1}^o$ ( $\omega _{t + 1}^o \leftarrow$ $\mathrm{LocalUpdate}$$(o,\omega _{t}^o)$\;
	 }
 ${\omega _{t + 1}} \leftarrow \frac{1}{{|\{O_v\}|}}\sum\nolimits_{o \in {O_v}} {\omega _{t + 1}^o}$\;
}
 $\mathrm{LocalUpdate}$$(o,\omega _{t}^o)$:\hfill $//$ Run on organization $o$ \;
 $\mathcal{B} \leftarrow$ (split $\mathcal{S}_o$ into batches of size $B$)\;
 \If{each local epoch $i$ from 1 to $E$}
 {\If{batch $b\in \mathcal{B}$}
 	{
 		$\omega  \leftarrow \omega  - \alpha  \cdot \nabla \mathcal{L}(\omega ;b)$\;
 	}
 }
\Return $\omega $ to cloud
		
\end{algorithm}
FedAVG algorithm is a critical mechanism in FedGRU to reduce the communication overhead in the process of transmitting parameters. This algorithm is an iterative process. For the $i$-th round of training, the models of the organizations participating in the training will be updated to the new global one.

\subsubsection{Federated Learning-based Gated Recurrent Unit neural network algorithm} FedGRU aims to achieve accurate and timely TFP through FL and GRU without compromising privacy. The overview of FedGRU is shown in Fig. \ref{fig-6}. It consists of four steps:
\begin{enumerate}[label=\roman*)]
	\item The cloud model is initialized through pre-training that utilizes domain-specific public datasets without privacy concerns;
	\item The cloud distributes the copy of the global model to all organizations, and each organization trains its copy on local data;
	\item Each organization uploads model updates to the cloud. The entire process does not share any private data, but instead sharing the encrypted parameters;
	\item The cloud aggregates the updated parameters uploaded by all organizations by the secure parameter aggregation mechanism to build a new global model, and then distributes the new global model to each organization.
\end{enumerate}
Given voluntary organization $\{ {O_v}\}  \subseteq \mathcal{O}$ and $o_v \in \{O_v\}$, referring to the GRU neural network in Section \ref{gru}, we have:
\begin{equation}\label{eq-7}
z_v^t = \sigma ({W^{({z_v})}} + {U^{({z_v})}}h_v^{t - 1}),
\end{equation}
\begin{equation}\label{eq-8}
r_v^t = \sigma ({W^{({r_v})}} + {U^{({r_v})}}h_v^{t - 1}),
\end{equation}
\begin{equation}\label{eq-9}
h{_v^{t\prime}}  = \tanh (Wx_v^t + r_v^t \odot Uh_v^{t - 1}),
\end{equation}
\begin{equation}\label{eq-10}
h_v^t = z_v^t \odot h_v^{t - 1} + (1 - z_v^t) \odot h{_v^{t^\prime} }.
\end{equation}
where $X = \{ x_v^1,x_v^2,\cdots,x_v^n\} ,Y = \{ y_v^1,y_v^2,\cdots,y_v^n\} ,H = \{ h_v^1,h_v^2,\cdots,h_v^n\} $ denote $o_v$'s input time series, $o_v$'s output time series and the hidden state of the cells, respectively. According to Eq. (\ref{eq-6}), the objective function of FedGRU is as follows:
\begin{equation}\label{eq-11}
\arg \mathop {\min }\limits_{\omega  \in {{\mathrm R}^d}} J(\omega ): = \sum\nolimits_{k = 1}^K {\frac{{\sum\nolimits_{i \in {D_v}} {{f_i}(\omega ) + \lambda h(\omega )} }}{D}}.
\end{equation}
The pseudocode of FedGRU is presented in Algorithm \ref{al-3}:
\begin{algorithm}[t]\label{al-2}
	\caption{Federated Learning-based Gated Recurrent Unit neural network (FedGRU) algorithm.}
	\LinesNumbered 
	\KwIn{$\{ {O_v}\}  \subseteq \mathcal{O}$, $X$, $Y$ and $H$. The mini-batch size $m$, the number of iterations $n$ and the learning rate $\alpha$. The optimizer $SGD$.}
	\KwOut{$J(\omega)$, $\omega $ and $W_v^r,W_v^z,W_v^h.$}
    According to  $X$, $Y$, $H$ and Eq. (\ref{eq-7})--(\ref{eq-11}), initialize the cloud model $J(\omega_{0})$, $\omega_{0}$, $W_v^{{r_0}}$, $W_v^{{z_0}}$, $W_v^{{h_0}}$, and $H_v^0$\;
    \ForEach{round $i = 1,2,3,\cdots$}
    {
    $\{O_v\} \leftarrow $ select volunteer from organizations to participate in this round of training\;
    \While{$g_\omega$ has not convergence}
    {
    	\ForEach{organization $o \in {O_v}$ \textbf{in parallel}}
    	{
    		 Conduct a mini-batch input time step $\{ {x_v}^{(i)}\} _{i = 1}^m$\;
    		 Conduct a mini-batch true traffic flow $\{ {y_v}^{(i)}\} _{i = 1}^m$\;
    		 Initalize $\omega _{t + 1}^o = \omega _t^o$\;
    		  ${g_\omega } \leftarrow {\nabla _\omega }\frac{1}{m}\sum\nolimits_{i = 1}^m {{{\big({f_\omega }(x_v^{(i)}) - y_v^{(i)}\big)}^2}} $\;
    		  $\omega _{t + 1}^o \leftarrow \omega _t^o + \alpha  \cdot SGD(\omega _t^o,{g_\omega })$\;
    		  Update the parameters $W_v^{{r_0}}$, $W_v^{{z_0}}$, $W_v^{{h_0}}$, and $H_v^0$\;
    		  Update reset gate $r$ and update gate $z$\;
    	}
    }
Collect the all parameters from $\{O_v\}$ to update ${\omega _{t + 1}}$. (Referring to the Algorithm \ref{al-1}.)\;
    }
\Return $J(\omega)$, $\omega $ and $W_v^r,W_v^z,W_v^h$
\end{algorithm}
\subsubsection{Joint-Announcement Protocol}\label{pro}
Generally, the number of participants in FedGRU is small. For example, WeBank worked with 7 auto insurance companies to build a traffic violation insurance pricing model using FL\footnote{\href{https://www.fedai.org/cases/a-case-of-traffic-violations-insurance-using-federated-learning/}{https://www.fedai.org/cases/a-case-of-traffic-violations-insurance-using-federated-learning/}}. In this case, since there are only 8 participants, we can define it as a small-scale federated learning model. However, a large number of participants may join FedGRU for traffic flow forecasting. When FedGRU is expanded to a large-scale scenario with many participants, FedAVG algorithm is hard to converge because of expensive communication overhead, thereby the accuracy of FedGRU will decrease. To address this issue, we design an improved FedAVG algorithm with a Joint-Announcement protocol in the aggregation mechanism to randomly select a certain proportion of organizations from a large number of participants in the $i$-th round training.

\begin{figure}[!t]
	\centering
	\includegraphics[width=1\linewidth]{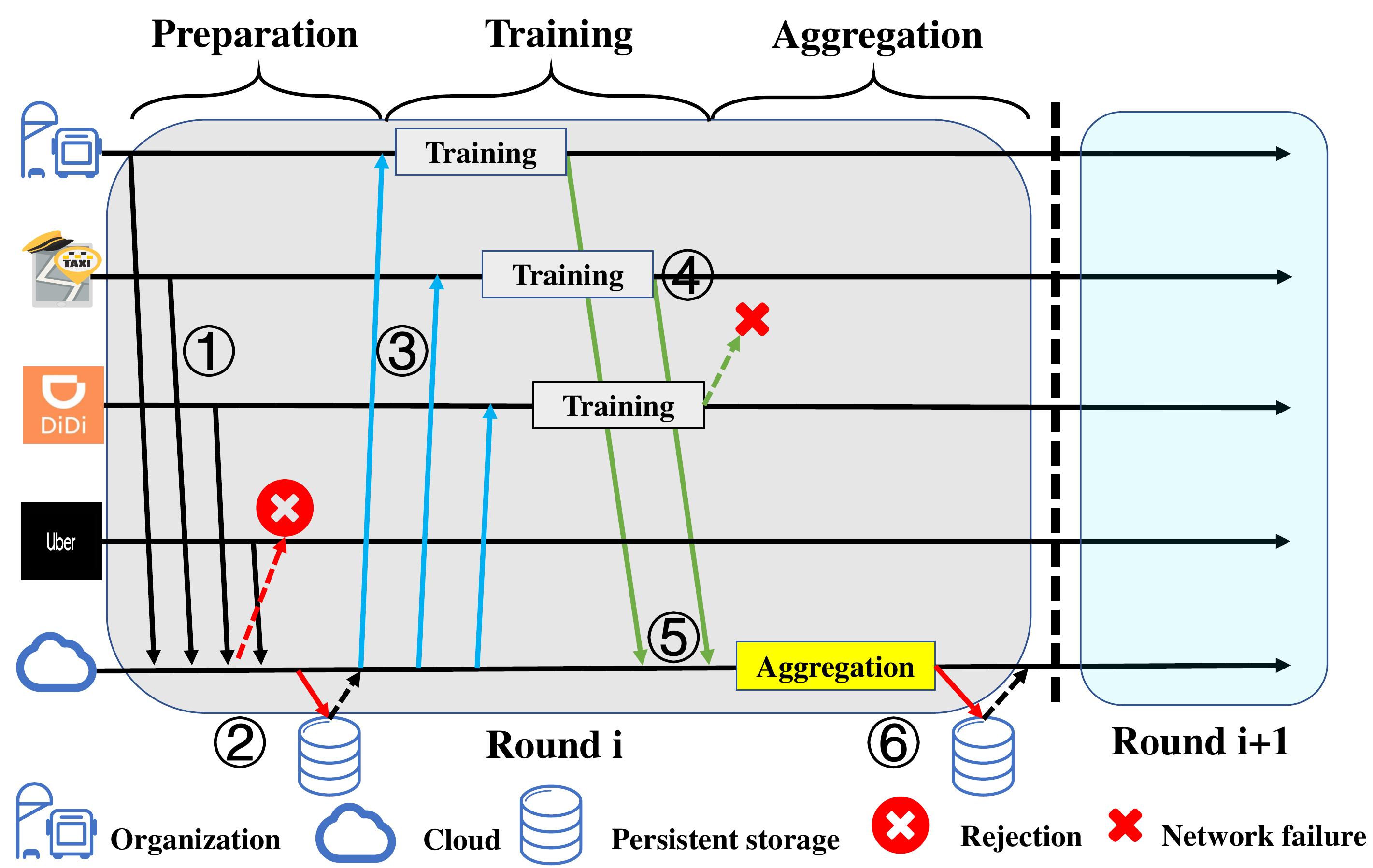}
	\caption{Federated learning joint-announcement protocol. }
	\label{fig-7}
\end{figure}

The participants in the Joint-Announcement protocol are organizations and the cloud, which is a cloud-based distributed service \cite{ref-19}. For $i$-th round of training, the protocol consists of three phases: preparation, training, and aggregation. The specific implementation phases of the protocol are given as follows:
\begin{enumerate}[label=\roman*)]
	\item \textit{Phase 1, Preparation:} Given a FL task (i.e., traffic flow prediction task in this paper), the organizations that voluntarily participate will check-in with the Cloud (as shown in Fig. \ref{fig-7}--\textcircled{1}). Who rejects ones represent if unwillingness to participate in this task or have other failures.
	\item  \textit{Phase 2, Training:} First, the cloud loads the pre-trained model (as shown in \textcircled{2}). Then the cloud sends the model checkpoint (i.e., gradient information) to the organizations (as shown in Fig. \ref{fig-7}--\textcircled{3}). The cloud randomly selects a fixed proportion (e.g., 10\%, 20\%, $\cdots$) of organizations to participate in this round of training (as shown in Fig. \ref{fig-7}--\textcircled{4}). Each organization will train the data locally and send the parameters to the cloud.
	\item \textit{Phase 3, Aggregation:} Subsequently, the cloud aggregates the parameters uploaded by organizations to update the global model through the security parameter aggregation mechanism (as shown in Fig. \ref{fig-7}--\textcircled{5}). In this  mechanism, the cloud executes FedAVG algorithm (presented in Section \ref{avg}) to reduce the uplink communication costs. The cloud updates the global model by sending checkpoints to persistent storage (as shown in Fig. \ref{fig-7}--\textcircled{6}). Finally, the global model sends update parameters to each organization.
\end{enumerate}

\subsection{Ensemble Clustering Federated Learning-Based Traffic Flow Prediction Algorithm}
\begin{figure}[!t]
	\centering
	\includegraphics[width=0.8\linewidth]{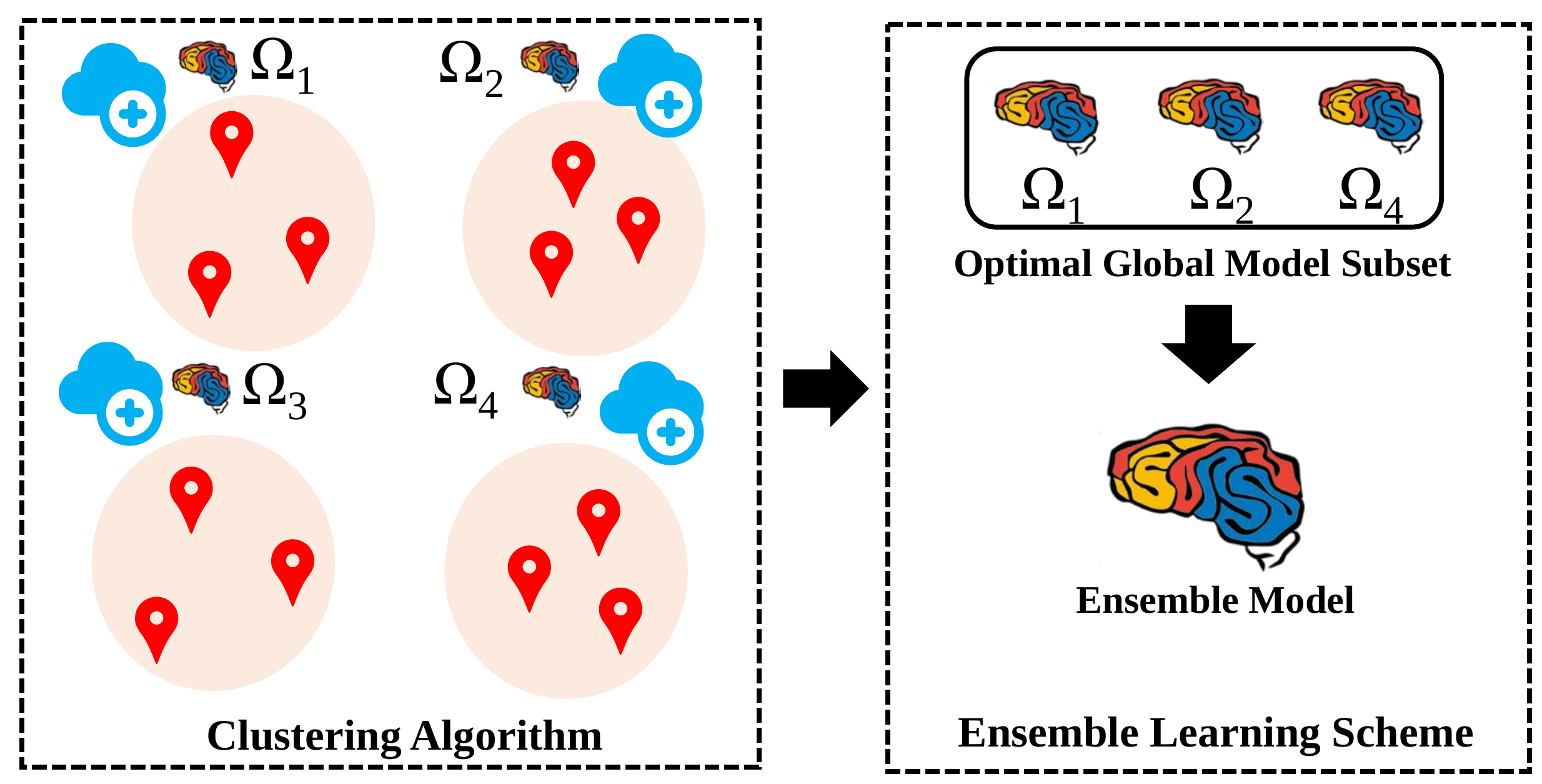}
	\caption{Ensemble clustering federated learning-based traffic flow prediction scheme. }
	\label{fig-18}
\end{figure}

Since the organizations select FL tasks based on its location information in the federated traffic flow prediction learning problem, for the same FL task, better spatio-temporal correlation of the data, leads to better performance. Based on the above hypothesis, we propose the ensemble clustering-based FedGRU algorithm to obtain better prediction accuracy and handle scenarios in which a large number of clients jointly work on training a traffic flow prediction problem by grouping organizations into $K$ clusters before implementing FedGRU. Then it integrates the global model of each cluster center by using an ensemble learning scheme, thereby obtains the best accuracy. In this scheme, the clustering decision is determined by using the latitude and longitude information of the organizations. We use the constrained K-Means algorithm proposed in \cite{ref-21}.  According to the definition of the constrained K-Means clustering algorithm, our goal is to determine the cluster center that minimizes the Sum of Squared Error (SSE). More formally, given a set $\mathcal{P}$ of $m$ points in $\mathbb{Re}^n$ (i.e organizations' location information) and the minimum cluster membership values ${\kappa _h}  \ge  0,h = 1,...,k$, cluster centers $C_1^t,C_2^t,...,C_k^t$ at iteration $t$, compute $C_1^{t + 1},C_2^{t + 1},...,C_k^{t + 1}$ at iteration $t+1$ in the following three steps as follows:
\begin{enumerate}[label=\roman*)]
	\item \textit{Sum of the Euclidean Metric Distance Squared:}
	\begin{equation}\label{eq-12}
	\begin{array}{l}
	d{(x,y)^2} = \sum\limits_{i = 1}^n {{{({x_i} - {y_i})}^2} = ||x - y||_2^2}, \\
	\mathrm{SSE} = \sum\nolimits_{i = 1}^m {\sum\nolimits_{h = 1}^k {{\tau _{i,h}}} } (\frac{1}{2}||{x^i} - C_h^t||_2^2).
	\end{array}
	\end{equation}
	where ${{\tau _{i,h}}}$ is a solution to the following linear program with $C_h^t$ fixed.
	\item \textit{Cluster Assignment:} To minimize \textit{SSE}, we have
	\begin{equation}\label{eq-13}
	\begin{array}{l}
	\mathop {\min }\limits_\tau  {\rm{  \qquad\qquad        }}\mathrm{SSE}\\
	{\rm{  \qquad\qquad \quad              }}\quad\sum\nolimits_{i = 1}^m {{\tau _{i,h}}}  \ge {\kappa _h},h = 1,2,\cdots,k\\
	\mathrm{subject}{\rm{ \quad \mathrm{to} }}\quad\sum\nolimits_{h = 1}^k {{\tau _{i,h}}}  = 1,i = 1,2,\cdots,m\\
	{\rm{    \qquad\qquad  \quad          }}\quad{\tau _{i,h}} \ge 0,i = 1,2,\cdots,m,h = 1,\cdots,k.{\rm{ }}
	\end{array}
	\end{equation}
	\item \textit{Cluster Update:} Update $C_h^{(t+1)}$ as follows:
	\begin{equation}\label{eq-14}
	{\rm{C}}_h^{t + 1} = \left\{ \begin{array}{l}
	\frac{{\sum\nolimits_{i = 1}^m {\tau _{i,h}^t{x^i}} }}{{\sum\nolimits_{i = 1}^m {\tau _{i,h}^t} }}\qquad \mathrm{if}\sum\nolimits_{i = 1}^m {\tau _{i,h}^t}  > 0,\\
	C_h^t\qquad \qquad \qquad \mathrm{otherwise}.
	\end{array} \right.
	\end{equation}
\end{enumerate}
If and only if SSE is minimum and $C_h^{t + 1} = C_h^t(h = 1,\cdots,k)$, we can obtain the optimal clustering center $C_k$ and the optimal set $\mathcal{P}_k =\{{P^i}\}_{i=1}^m$. Let ${\Omega}_k = \{\omega_1, \omega_2, \cdots, \omega_k\}$ denote the global model of the optimal set $\mathcal{P}_k$. As shown in Fig. \ref{fig-18}, we utilize the ensemble learning scheme to find the optimal ensemble model by integrating the global model from $\Omega_k$ with the best accuracy after executing the constrained K-Means algorithm. More formally, such a scheme needs to find an optimal global model subset of the following equation:
\begin{equation}\label{optiaml}
\mathop {\max }\limits_\Omega  \frac{1}{{|{\Omega _j}|}}\sum\limits_{j = 1}^{j \le k} {{\Omega _j}} ,\mathrm{where}\;{\Omega _j} \subseteq {\Omega _k}.
\end{equation}

The ensemble clustering-based FedGRU is thus presented in Algorithm \ref{al-3}. It consists of three steps:
\begin{enumerate}[label=\roman*)]
	\item Given organization set $\mathcal{O}$, we random initialize cluster centers $C_h^t$, and execute the constrained K-Means algorithm;
	\item With the optimal clustering center $C_k$ and the optimal set ${O_k} =\{{O^i}\}_{i=1}^m$, the cloud executes the ensemble scheme to find the optimal global model set $\Omega_j$ (i.e., subset of $\Omega_k$);
	\item The cloud sends the new global model to each organization.
\end{enumerate}
\begin{algorithm}[t]\label{al-3}
	\caption{Ensemble clustering federated learning-based FedGRU algorithm.}
	\LinesNumbered 
	\KwIn{Organizations set ${\mathcal{O}} = \{ {O^i}\} _{i = 1}^m$.}
	\KwOut{$J(\omega)$, $\omega $ and $W_v^r,W_v^z,W_v^h.$}
	Initialize random cluster center $C_h^t$\;
	\While{$C_h^{t + 1} \ne C_h^t(h = 1,\cdots,k)$}
	{
		Execute the constrained K-Means algorithm's step 1 and step 2 (Referring to Eq. (\ref{eq-12})--(\ref{eq-13}))\;
		Update $C_h^{(t+1)}$ according to Eq. (\ref{eq-14}) in step 3 of the constrained K-Means algorithm\;
	}
\ForEach{clustering center $C_k$ and the optimal set ${\mathcal{{O}}}_k =\{{O^i}\}_{i=1}^k$}
{
Execute FedGRU algorithm\;
}
Obtain the global model set $\Omega_k$\;
Execute the ensemble learning scheme to find the optimal global model subset $\Omega_j$ (Referring to Eq. (\ref{optiaml})\;
The cloud sends the new global model to each organization\;
\Return $J(\omega)$, $\omega $ and $W_v^r,W_v^z,W_v^h$.
\end{algorithm}

\section{EXPERIMENTS}\label{sec-4}
In this experiment, the proposed FedGRU and clustering-based FedGRU algorithms are applied to the real-world data collected from the Caltrans Performance Measurement System (PeMS) \cite{ref-13} database for performance demonstration. The traffic flow data in PeMS database was collected from over 39,000 individual detectors  in real time. These sensors span the freeway system across all major metropolitan areas of the State of California \cite{ref-1}. In this paper, traffic flow data collected during the first three months of 2013 is used for experiments. We select the traffic flow data in the first two months as the training dataset and the third month as the testing dataset. Furthermore, since the traffic flow data is time-series data, we need to use them at the previous time interval, i.e., ${x_{t - 1}},{x_{t - 2}},\cdots,{x_{t - r}}$, to predict the traffic flow at time interval $t$, where $r$ is the length of the history data window.

We adopt Mean Absolute Error (MAE), Mean Square Error (MSE), Root Mean Square Error (RMSE), and Mean Absolute Percentage Error (MAPE) to indicate the prediction accuracy as follows:

\begin{equation}
\mathrm{MAE} = \frac{1}{n}\sum\limits_{i = 1}^n {|y_i - } \hat{y}_p|,
\end{equation}
\begin{equation}
\mathrm{MSE} = \frac{1}{n}\sum\limits_{i = 1}^n {({y_i} - \hat{y}_p} {)^2},
\end{equation}
\begin{equation}
\mathrm{RMSE} = [\frac{1}{n}\sum\limits_{i = 1}^n {{{(|y_i - \hat{y}_p|)}^2}{]^{\frac{1}{2}}}},
\end{equation}
\begin{equation}
\mathrm{MAPE} = \frac{{100\% }}{n}\sum\limits_{i = 1}^n {|\frac{{\hat{y}_p - {y_i}}}{{{y_i}}}|}. 
\end{equation}
Where $y_i$ is the observed traffic flow, and $\hat{y}_p$ is the predicted traffic flow.

Without loss of generality, we assume that the detector stations are distributed and independent and the data cannot be exchanged arbitrarily among them. In the secure parameter aggregation mechanism, PySyft \cite{ref-23} framework is adopted to encrypt the parameters\footnote{\href{https://github.com/OpenMined/PySyft}{https://github.com/OpenMined/PySyft}}.  The FedGRU code is available at \href{https://github.com/niklausliu/TF_FedGRU_demo}{https://github.com/niklausliu/TF\_FedGRU\_demo}.

For the cloud and each  organization, we use mini-batch SGD for model optimization. PeMS dataset is split equally and assigned to 20 organizations. During the simulation, learning rate $\alpha =0.001$, mini-batch size $m=128$, and $|O_v|=20$. Note that the client $C = 2$ of the FedGRU model is the default setting in FL \cite{ref-16}. All experiments are conducted using TensorFlow and PyTorch \cite{pytorch} with Ubuntu 18.04. 
\subsection{FedGRU Model Architecture} In the context of deep learning, proper hyperparameter selection, e.g. the size of the input layer, the number of hidden layers, and hidden units in each hidden layer, is a notable factor that determines the model performance. In this section, we investigate the performance of FedGRU with different hyperparameter configurations and try to determine the best-performing neural network architecture for it. Additionally, we also obtain the optimal length of history data window $r=12$. In particular, we employ $r=12$, the number of hidden layers in $[1,3]$, and the number of hidden units in $\{50, 100, 150\}$ \cite{ref-1} to adjust the structure of FedGRU. Furthermore, we utilize the grid search approach to find the best architecture for FedGRU.

We first evaluate the performance of FedGRU on a 5-min traffic flow prediction task through MAE, MSE, RMSE, and MAPE. After performing the grid search, we obtain the best architecture of FedGRU as shown in Table \ref{tb-1}. The optimal architecture consists of two hidden layers, each with a hidden layer number of $\{50, 50\}$. The results show that the optimal number of hidden layers in our experiment is two. From a model perspective, the number of hidden layers of FedGRU model should not be too small or too large. Our results confirm these facts.
\begin{table*}[t]
	\centering
	\caption{Structure of FedGRU For Traffic Flow Prediction}
	\begin{tabular}{cccccccc}
		\toprule
		Metrics & Time steps  &Hidden layers & Hidden units & MAE &MSE &RMSE &MAPE\\
		\midrule
		\multirow{9}*{FedGRU (default setting)} & \multirow{9}*{5} & \multirow{3}*{1} & 50 &9.03&103.24&13.26&19.12 \\
		~ &        ~ &  ~ & 100 &8.96&102.36&14.32&18.94\\
		~&~&~&150&8.46&101.05&14.98&18.46\\
		\cline{3-8}
		~ &        ~ &  \multirow{3}*{2}& 50, 50  &\textbf{7.96} &\textbf{101.49}  &\textbf{11.04}   &\textbf{17.82}\\
		~&~&~&100, 100&8.64&102.21&15.06&19.22\\
		~&~&~&150, 150&8.75&102.91&14.93&19.35\\
		\cline{3-8}
		~ & ~ &  \multirow{3}*{3}& 50, 50, 50 &8.29&102.17&12.05&18.62\\
		~&~&~&100, 100, 100&8.41&103.01&13.45&18.96\\
		~&~&~&150, 150, 150&8.75&103.98&13.74&19.24\\	
		\bottomrule
	\end{tabular}
\label{tb-1}
\end{table*}
\begin{table*}[!t]
	\centering
	\caption{Performance Comparsion of MAE, MSE, RMSE, And MAPE For FedGRU, LSTM, SAE, And SVM}
	\begin{tabular}{ccccc} 
		\toprule
		Metrics  &MAE	&MSE	&RMSE	&MAPE		\\
		\midrule
		FedGRU (default setting) &7.96 &101.49  &11.04   &17.82\%    \\
		GRU \cite{ref-15}      &7.20	&99.32	&9.97    &17.78\%		\\
		SAE \cite{ref-1}     &8.26	&99.82	&11.60    &19.80\%		\\
		LSTM \cite{ref-32}      &8.28	&107.16	&11.45    &20.32\%		\\
		SVM  \cite{ref-54}     &8.68	&115.52	&13.24    &22.73\%		\\
		\bottomrule
	\end{tabular}
	\label{tb-2}
\end{table*}
\subsection{Traffic Flow Prediction Accuracy}\label{sec-b} We compared the performance of the proposed FedGRU model with that of GRU, SAE, LSTM, and support vector machine (SVM) with an identical simulation configuration. Among these five competing methods, FedGRU is a federated machine learning model, and the rest are centralized ones. Among them, GRU is a widely-adopted baseline model that has good performance for traffic flow forecast tasks, as aforementioned in Section \ref{sec-3}, and SVM is a popular machine learning model for general prediction applications \cite{ref-1}. In all investigations, we use the same PeMS dataset. The prediction results are given in Table \ref{tb-2} for 5-min ahead traffic flow prediction. From the simulation results, it can be observed that MAE of FedGRU is lower than those of SAEs, LSTM, and SVM but higher than that of GRU. Specifically, MAE of FedGRU is 9.04\% lower than that of the worst case (i.e., SVM) in this experiment. This result is contributed by the fact that FedGRU inherits the advantages of GRU's outstanding performance in prediction tasks. 

\begin{figure}[t]
	\centering
	\large
	\subfigure []{\includegraphics[width=0.7\linewidth]{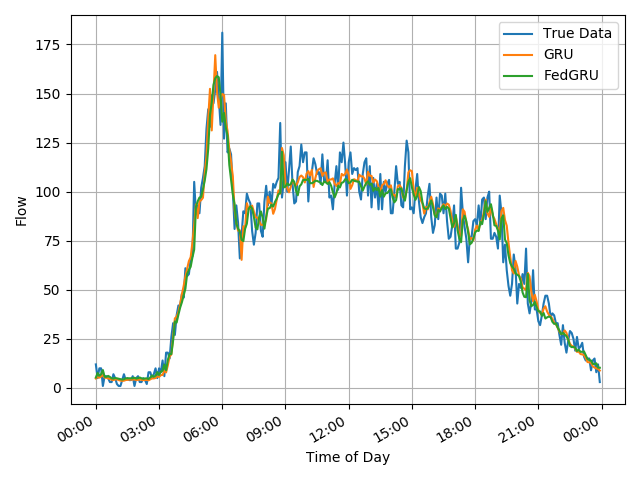}
		\label{a-1}}
	\hfill
	\subfigure[]{	\includegraphics[width=0.7\linewidth]{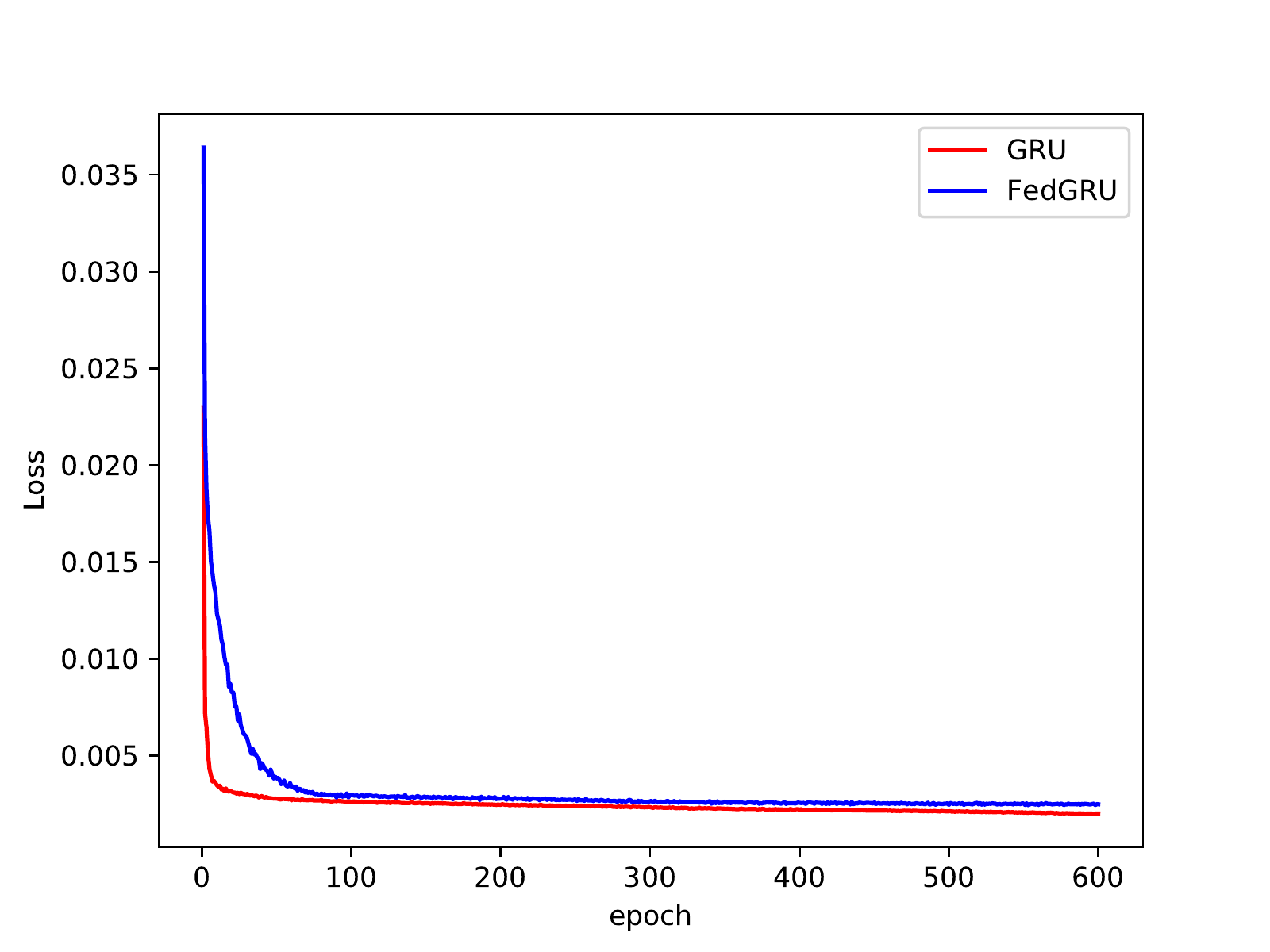}
		\label{b-1}}
	\caption{(a) Traffic flow prediction of GRU model and FedGRU model. (b) Loss of GRU model and FedGRU model.}
	\label{fig-10}
\end{figure}

Fig. \ref{a-1} shows a comparison between GRU and FedGRU for a 5-min traffic flow prediction task. We can find that the predict results of FedGRU model are very close to that of GRU. This is because the core technique of FedGRU to prediction is GRU structure, so the performance of FedGRU is comparable to GRU model. Furthermore, FedGRU can protect data privacy by keeping the training dataset locally. Fig. \ref{b-1} illustrates the loss of GRU model and FedGRU model. From the results, the loss of FedGRU model is not significantly different from GRU model. This proves that FedGRU model has good convergence and stability. In a word, FedGRU can achieve accurate and timely traffic flow prediction without compromising privacy. 

\subsection{Performance Comparison of FedGRU Model Under Different Client Numbers}\label{ex-2}
\begin{figure}[t]
	\centering
	\includegraphics[width=0.7\linewidth]{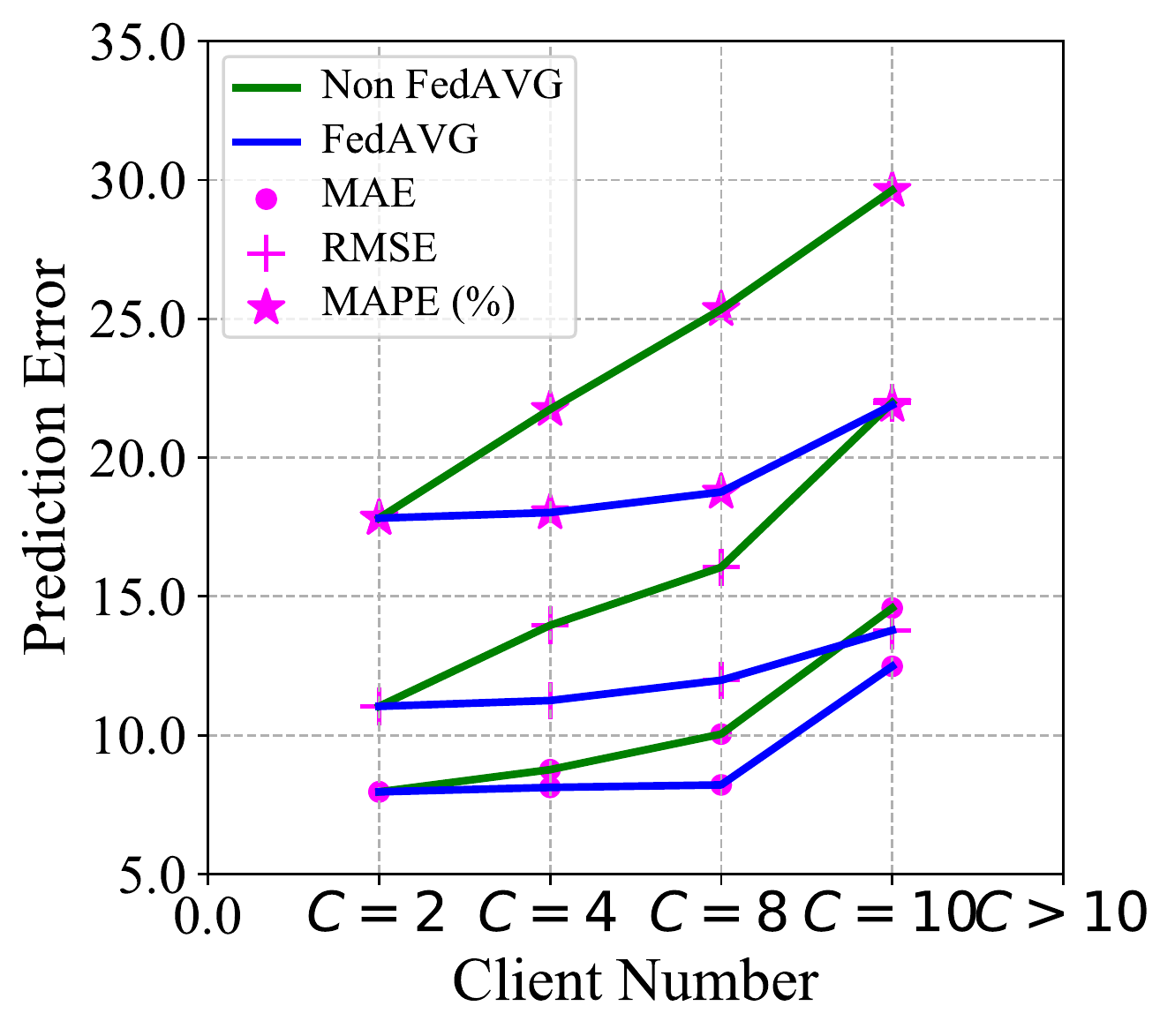}
	\caption{The prediction error of FedGRU model with different client numbers. }
	\label{fig-12}
\end{figure}

In Section \ref{sec-b}, the default client number is set $C = 2$. However, it is highly plausible that traffic data can be gathered by more than two entities, e.g., organizations and companies. In this experiment, we explore the impact of different client numbers (i.e., $C=2, 4, 8, 10$) on the performance of FedGRU. The simulation results are presented in Fig. \ref{fig-12}, where we observe  that the number of clients has an adverse influence on the performance of FedGRU. The reason is that more clients introduce increasing communication overhead to the underlying communication infrastructure, which makes it more difficult for the cloud to simultaneously perform aggregation of gradient information. Furthermore, such overhead may cause communication failures in some clients, causing clients to fail to upload gradient information, thereby reducing the accuracy of the global model. 

In this paper, we initially use FedAVG algorithm to alleviate the expensive communication overhead issue. FedAVG reduces communication overhead by i) computing the average gradient of a batch size samples on the client and ii) computing the average aggregation gradient from all clients. Fig. \ref{fig-12} shows that FedAVG performs well when the number of clients is less than 8, but when the number of clients exceeds 8, the performance of FedAVG starts to decline. The reason is that, when the number of clients exceeds a certain threshold (e.g., $C = 8$), the probability of client failure will increase, which causes FedAVG to calculate wrong gradient information. Nevertheless, FedAVG is significant for reducing communication overhead because the number of entities involved in predicting traffic flow tasks in real life is usually small. Therefore, we need to propose a new communication protocol for large-scale organizations to solve the problem of communication overhead.
\begin{figure}[!t]
	\centering
	\includegraphics[width=0.7\linewidth]{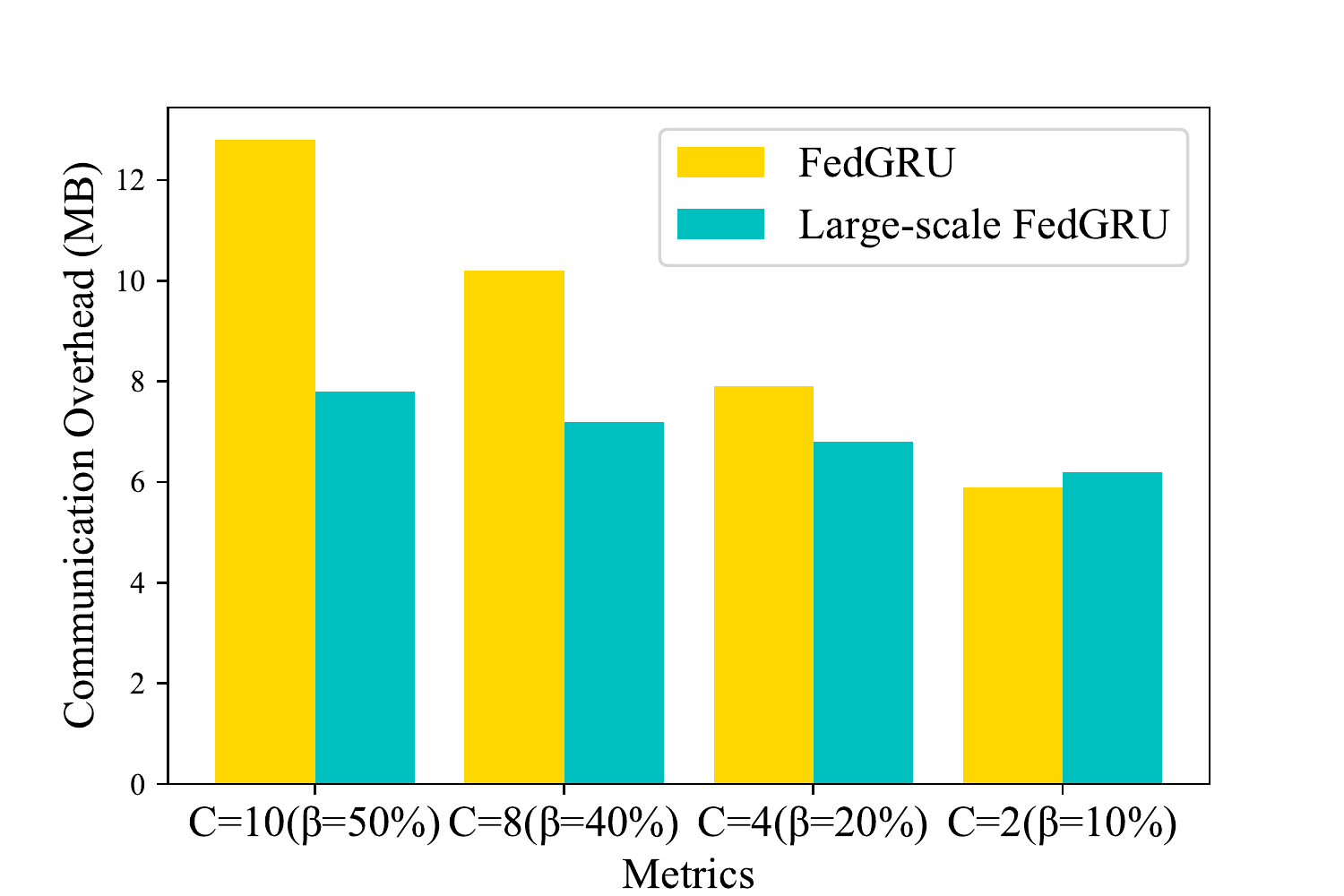}
	\caption{Communication overhead between large-scale FedGRU and FedGRU. }
	\label{fig-11}
\end{figure}
\begin{figure}[!t]
	\centering
	\includegraphics[width=0.7\linewidth]{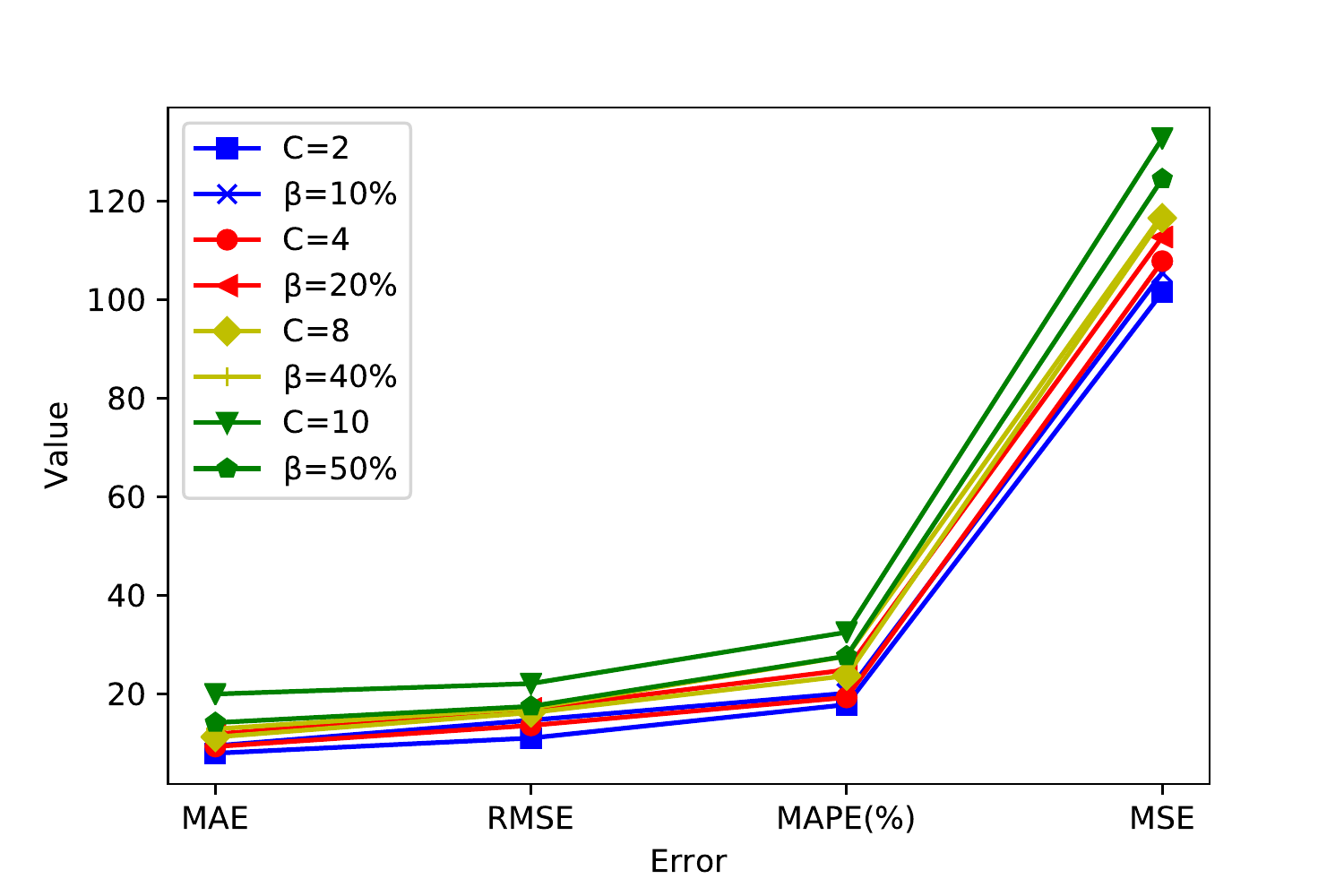}
	\caption{The prediction results of models with different participation ratios ($\beta  = 10\% ,20\% ,40\% ,50\% $). }
	\label{fig-15}
\end{figure}
\begin{figure*}[!t]
	\centering
	\includegraphics[width=0.8\linewidth]{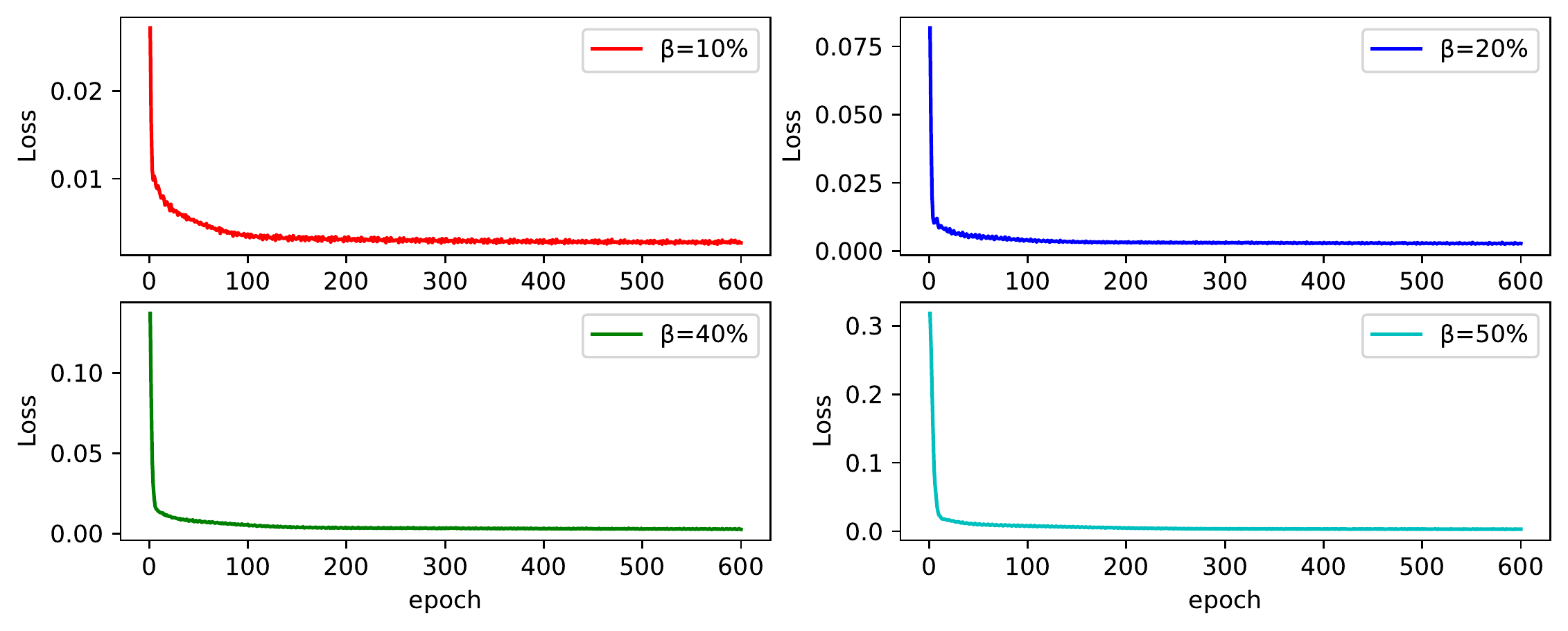}
	\caption{Loss of FedGRU model with different participation ratios. ($\beta  = 10\% ,20\% ,40\% ,50\% $)}
	\label{fig-16}
\end{figure*}
\subsection{Traffic Flow Prediction With Large-scale FedGRU Model}
In Section \ref{ex-2}, the experimental results show that  FedAVG is no longer suitable for large-scale organizations when $C \ge 8$. However, in real life, we sometimes cannot avoid large-scale organization participation in FedGRU model. To solve this problem, we design the joint-announcement protocol, which can randomly select a certain proportion of organizations from a large number of participating organizations to participate in the $i$-th round training. In this experiment, we set the participation ratio $\beta \in \{10\% ,20\% ,40\% ,50\%\} $ and set $C=20$. Then we compare these four cases with the ones of section \ref{ex-2}. 

In this experiment, we first focus on the communication overhead of a large-scale FedGRU model. In Fig. \ref{fig-11}, we show that FedGRU with joint-announcement protocol can significantly reduce the communication overhead. Specifically, when $C = 10 ~ (\beta = 50\%)$, the communication overhead of FedGRU with the joint-announcement protocol is reduced by 64.10\% compared to FedGRU with FedAVG algorithm. The Joint-announcement protocol first performs sub-sampling on participating organizations, which can reduce the number of participants. Then it uses FedAVG algorithm to calculate the average gradient information, which guarantees the reliability of model training. Furthermore, experimental results show that such a protocol is robust to the number of participants, that is, the performance of the protocol is not affected by the number of participants.

Fig. \ref{fig-15} shows the prediction results of models with different participation ratios. It shows that when $C=10 ~  (\beta = 50\%)$, the prediction results of large-scale FedGRU is the most different from FedGRU. When $C=10  ~(\beta = 50\%)$, MAE of large-scale FedGRU is 29.08\% upper than MAE of FedGRU. This is because the performance of FedAVG starts to decrease when $C\ge 8$ (as shown in Fig. \ref{fig-15}), and FedGRU with joint-announcement protocol can control the number of participants through sub-sampling to maintain the performance of FedAVG. In Fig. \ref{fig-16}, we can find the loss of models with different $\beta$. It shows that the larger $\beta$ of the model, the greater the loss in the early training period. But $\beta$ does not affect the convergence of the models. Therefore, FedGRU using joint-announcement protocol can maintain good stability, robustness and efficiency.

\subsection{Traffic Flow Prediction With Ensemble Clustering-Based FedGRU Model}\label{ex-4}

In this subsection, we evaluate an ensemble clustering-based FedGRU in scenarios where a large number of clients jointly work on training a traffic flow prediction problem. In particular, we examine the effect of $K$ on the proposed ensemble clustering-based mechanism. In Table \ref{tb-3}, we show the accuracy of Clustering-Based FedGRU model when the cluster centers $K = 0, 2, 4, 8, 10$. The results indicate that the proposed ensemble clustering-Based FedGRU model achieves the best prediction accuracy and can further improve the performance of the original FedGRU model. Compared with GRU model, the Clustering-Based FedGRU model can even outperform the centralized GRU model when $K = 8, 10$, which still compromises the data privacy. The reason is that the ensemble clustering-based scheme can improve the prediction accuracy by classifying similar spatio-temporal features into the same cluster and integrating the advantages of the optimal global model set. Furthermore, in such a scheme, it is easy for FedGRU to find the optimal global model subset because $K$ is relatively small. Therefore, our proposed ensemble clustering-based federated learning scheme can further improve the accuracy of prediction, thereby achieving accurate and timely traffic flow prediction.
\begin{table}[!t]
	\centering
	\caption{Performance comparison of FedGRU algorithm and Clustering-Based FedGRU algorithm}
	\begin{tabular}{ccccc} 
		\toprule
		Metrics       &MAE	&MSE	&RMSE	&MAPE		\\
		\midrule
		FedGRU        &7.96 &101.49  &11.04   &17.82\%    \\
		FedGRU ($K=2$)&7.89	&100.98	&10.82    &17.16\%		\\
		FedGRU ($K=4$)&7.42	&99.98	&10.01    &16.85\%		\\
		FedGRU ($K=8$)&\textbf{7.17}	&\textbf{99.16}	&\textbf{9.86}    &\textbf{16.22\%}		\\
		FedGRU ($K=10$)&\textbf{6.85}	&\textbf{97.77}	&\textbf{9.49}    &\textbf{14.69\%}		\\
		\bottomrule
	\end{tabular}
	\label{tb-3}
\end{table}

\subsection{Discussion} \label{diss}
In this subsection, we further discuss the advantages and limitations of FedGRU in different scenarios for predicting traffic flow.
In the previous subsections, we carry out a series of comprehensive case studies to show the effectiveness of our proposed method. Based on the above empirical results, the following observations can be derived:
\begin{enumerate}[label=\roman*)]
	\item Communication overhead is the bottleneck of FedGRU model. For a large-scale FedGRU model, the joint-announcement protocol helps solve the communication overhead problem. It mitigates the communication overhead of FedGRU model by reducing the number of participants participating in each round of communication. 
	\item Global model updates for the client in FedGRU model are not synchronized. For example, FedGRU may fail to synchronize global model updates due to the failure of some clients. This may potentially make the local model of arbitrary clients deviate from the current global one, which affects the next global model update. To address this problem, we use random sub-sampling to select organizations that participate in the $i$-th round of training, as reducing the number of participants participating in the $i$-th training can reduce the probability of client failure, thereby alleviating the out-of-sync issue.
	\item Statistical heterogeneity issue is a problem that needs to be solved. Due to a large number of organizations involved in training, the large-scale FedGRU model faces a challenge: the local data are not i.i.d.\ \cite{ref-47,ref-52}. Organizations often generate and collect data across the network in a completely different way \cite{ref-48}. This data generation paradigm violates the i.i.d.\ assumption commonly used in distributed optimization, increasing the likelihood of sprawl and possibly increasing the complexity of modeling, analysis, and evaluation \cite{ref-49}.
\end{enumerate}
\subsection{Privacy Analysis}
According to the definition of information-based privacy, we discuss the privacy protection capabilities of the proposed FedGRU model from the following aspects:
\begin{enumerate}[label=\roman*)]
\item \textbf{Data Access:} The proposed model is developed based on the federated learning framework, and its core idea is a distributed privacy protection framework. Specifically, the proposed model achieves accurate traffic flow prediction by aggregating encrypted parameters rather than accessing the original data, which guarantees the model's privacy protection for the data.
\item \textbf{Model Performance:} Experimental results show that the performance of the proposed model is comparable to GRU model. GRU model is a centralized machine learning model, which needs to aggregate a large amount of raw data to achieve high-precision traffic flow prediction. Furthermore, there is a trade-off between traffic flow prediction and privacy. The proposed model achieves comparable results to a centralized machine learning approach under the constraint of privacy preservation, therefore demonstrates its superiority.
\end{enumerate}

\section{Conclusion}\label{sec-6}
In this paper, we propose a FedGRU algorithm for traffic flow prediction with federated learning for privacy preservation. FedGRU does not directly access distributed organizational data but instead employs secure parameter aggregation mechanism to train a global model in a distributed manner. It aggregates the gradient information uploaded by all locally trained models in the cloud to construct the global one for traffic flow forecast. We evaluate the performance of FedGRU on a PeMS dataset and compared it with GRU, LSTM, SAE, and SVM, which all potentially compromise user privacy during the forecast. The results show that the proposed method performs comparably to the competing methods with minuscule accuracy degradation with privacy well-preserved.
Furthermore, we apply an ensemble clustering-based FedGRU for TFP to further improve the model performance. We also demonstrate by empirical studies that the proposed joint-announcement protocol is efficient in reducing the communication overhead for FedGRU by 64.10\% compared with centralized models.

To the best of our knowledge, this is among the pioneering work for traffic flow forecasts with federated deep learning. In the future, we plan to apply Graph Convolutional Network (GCN) \cite{ref-44} to the federated learning framework to better capture the spatial-temporal dependency among traffic flow data to further improve the prediction accuracy.

\ifCLASSOPTIONcaptionsoff
  \newpage
\fi



%
\bibliographystyle{IEEEtran}
\bibliography{reference}

\begin{thebibliography}{10}
\providecommand{\url}[1]{#1}
\csname url@samestyle\endcsname
\providecommand{\newblock}{\relax}
\providecommand{\bibinfo}[2]{#2}
\providecommand{\BIBentrySTDinterwordspacing}{\spaceskip=0pt\relax}
\providecommand{\BIBentryALTinterwordstretchfactor}{4}
\providecommand{\BIBentryALTinterwordspacing}{\spaceskip=\fontdimen2\font plus
\BIBentryALTinterwordstretchfactor\fontdimen3\font minus
  \fontdimen4\font\relax}
\providecommand{\BIBforeignlanguage}[2]{{%
\expandafter\ifx\csname l@#1\endcsname\relax
\typeout{** WARNING: IEEEtran.bst: No hyphenation pattern has been}%
\typeout{** loaded for the language `#1'. Using the pattern for}%
\typeout{** the default language instead.}%
\else
\language=\csname l@#1\endcsname
\fi
#2}}
\providecommand{\BIBdecl}{\relax}
\BIBdecl

\bibitem{ref-1}
Y.~Lv, Y.~Duan, W.~Kang, Z.~Li, and F.-Y. Wang, ``Traffic flow prediction with
  big data: a deep learning approach,'' \emph{IEEE Transactions on Intelligent
  Transportation Systems}, vol.~16, no.~2, pp. 865--873, 2014.

\bibitem{ref-2}
N.~Zhang, F.-Y. Wang, F.~Zhu, D.~Zhao, S.~Tang \emph{et~al.}, ``Dynacas:
  Computational experiments and decision support for {ITS},'' 2008.

\bibitem{ref-66}
K.~{Ren}, Q.~{Wang}, C.~{Wang}, Z.~{Qin}, and X.~{Lin}, ``The security of
  autonomous driving: Threats, defenses, and future directions,''
  \emph{Proceedings of the IEEE}, vol. 108, no.~2, pp. 357--372, Feb 2020.

\bibitem{ref-67}
X.~{Lin}, X.~{Sun}, P.~{Ho}, and X.~{Shen}, ``Gsis: A secure and
  privacy-preserving protocol for vehicular communications,'' \emph{IEEE
  Transactions on Vehicular Technology}, vol.~56, no.~6, pp. 3442--3456, Nov
  2007.

\bibitem{ref-64}
Y.~{Zhang}, C.~{Xu}, H.~{Li}, K.~{Yang}, N.~{Cheng}, and X.~S. {Shen},
  ``Protect: Efficient password-based threshold single-sign-on authentication
  for mobile users against perpetual leakage,'' \emph{IEEE Transactions on
  Mobile Computing}, pp. 1--1, 2020.

\bibitem{ref-61}
R.~N. Fries, M.~R. Gahrooei, M.~Chowdhury, and A.~J. Conway, ``Meeting privacy
  challenges while advancing intelligent transportation systems,''
  \emph{Transportation Research Part C: Emerging Technologies}, vol.~25, pp.
  34--45, 2012.

\bibitem{ref-63}
C.~{Zhang}, J.~J.~Q. {Yu}, and Y.~{Liu}, ``Spatial-temporal graph attention
  networks: A deep learning approach for traffic forecasting,'' \emph{IEEE
  Access}, vol.~7, pp. 166\,246--166\,256, 2019.

\bibitem{ref-7}
S.~Madan and P.~Goswami, ``A novel technique for privacy preservation using
  k-anonymization and nature inspired optimization algorithms,''
  \emph{Available at SSRN 3357276}, 2019.

\bibitem{ref-43}
J.~L. {Ny}, A.~{Touati}, and G.~J. {Pappas}, ``Real-time privacy-preserving
  model-based estimation of traffic flows,'' in \emph{2014 ACM/IEEE
  International Conference on Cyber-Physical Systems (ICCPS)}, April 2014, pp.
  92--102.

\bibitem{ref-16}
J.~Kone{\v{c}}n{\`y}, H.~B. McMahan, F.~X. Yu, P.~Richt{\'a}rik, A.~T. Suresh,
  and D.~Bacon, ``Federated learning: Strategies for improving communication
  efficiency,'' \emph{arXiv preprint arXiv:1610.05492}, 2016.

\bibitem{ref-11}
X.~Yuan, X.~Wang, C.~Wang, J.~Weng, and K.~Ren, ``Enabling secure and fast
  indexing for privacy-assured healthcare monitoring via compressive sensing,''
  \emph{IEEE Transactions on Multimedia (TMM)}, vol.~18, no.~10, pp. 1--13,
  2016.

\bibitem{ref-24}
M.~S. Ahmed, ``Analysis of freeway traffic time series data and their
  application to incident detection,'' \emph{Equine Veterinary Education},
  vol.~6, no.~1, pp. 32--35, 1979.

\bibitem{ref-25}
M.~Van Der~Voort, M.~Dougherty, and S.~Watson, ``Combining kohonen maps with
  arima time series models to forecast traffic flow,'' \emph{Transportation
  Research Part C: Emerging Technologies}, vol.~4, no.~5, pp. 307--318, 1996.

\bibitem{ref-26}
S.~Lee and D.~B. Fambro, ``Application of subset autoregressive integrated
  moving average model for short-term freeway traffic volume forecasting,''
  \emph{Transportation Research Record}, vol. 1678, no.~1, pp. 179--188, 1999.

\bibitem{ref-27}
B.~M. Williams and L.~A. Hoel, ``Modeling and forecasting vehicular traffic
  flow as a seasonal arima process: Theoretical basis and empirical results,''
  \emph{Journal of transportation engineering}, vol. 129, no.~6, pp. 664--672,
  2003.

\bibitem{ref-28}
J.~J.~Q. {Yu}, A.~Y.~S. {Lam}, D.~J. {Hill}, Y.~{Hou}, and V.~O.~K. {Li},
  ``Delay aware power system synchrophasor recovery and prediction framework,''
  \emph{IEEE Transactions on Smart Grid}, vol.~10, no.~4, pp. 3732--3742, July
  2019.

\bibitem{ref-29}
G.~A. Davis and N.~L. Nihan, ``Nonparametric regression and short-term freeway
  traffic forecasting,'' \emph{Journal of Transportation Engineering}, vol.
  117, no.~2, pp. 178--188, 1991.

\bibitem{ref-30}
C.-C. Chang and C.-J. Lin, ``Libsvm: A library for support vector machines,''
  \emph{ACM transactions on intelligent systems and technology (TIST)}, vol.~2,
  no.~3, p.~27, 2011.

\bibitem{ref-31}
D.~Svozil, V.~Kvasnicka, and J.~Pospichal, ``Introduction to multi-layer
  feed-forward neural networks,'' \emph{Chemometrics and intelligent laboratory
  systems}, vol.~39, no.~1, pp. 43--62, 1997.

\bibitem{ref-32}
X.~Ma, Z.~Tao, Y.~Wang, H.~Yu, and Y.~Wang, ``Long short-term memory neural
  network for traffic speed prediction using remote microwave sensor data,''
  \emph{Transportation Research Part C: Emerging Technologies}, vol.~54, pp.
  187--197, 2015.

\bibitem{ref-33}
Y.~Tian and L.~Pan, ``Predicting short-term traffic flow by long short-term
  memory recurrent neural network,'' in \emph{2015 IEEE international
  conference on smart city/SocialCom/SustainCom (SmartCity)}.\hskip 1em plus
  0.5em minus 0.4em\relax IEEE, 2015, pp. 153--158.

\bibitem{ref-15}
R.~{Fu}, Z.~{Zhang}, and L.~{Li}, ``Using lstm and gru neural network methods
  for traffic flow prediction,'' in \emph{2016 31st Youth Academic Annual
  Conference of Chinese Association of Automation (YAC)}, Nov 2016, pp.
  324--328.

\bibitem{ref-35}
J.~J.~Q. {Yu}, W.~{Yu}, and J.~{Gu}, ``Online vehicle routing with neural
  combinatorial optimization and deep reinforcement learning,'' \emph{IEEE
  Transactions on Intelligent Transportation Systems}, vol.~20, no.~10, pp.
  3806--3817, Oct 2019.

\bibitem{ref-34}
J.~J.~Q. {Yu} and J.~{Gu}, ``Real-time traffic speed estimation with graph
  convolutional generative autoencoder,'' \emph{IEEE Transactions on
  Intelligent Transportation Systems}, vol.~20, no.~10, pp. 3940--3951, Oct
  2019.

\bibitem{ref-36}
U.~M. A{\"\i}vodji, S.~Gambs, M.-J. Huguet, and M.-O. Killijian, ``Meeting
  points in ridesharing: A privacy-preserving approach,'' \emph{Transportation
  Research Part C: Emerging Technologies}, vol.~72, pp. 239--253, 2016.

\bibitem{ref-37}
B.~Y. He and J.~Y. Chow, ``Optimal privacy control for transport network data
  sharing,'' \emph{Transportation Research Part C: Emerging Technologies},
  2019.

\bibitem{ref-55}
Y.~Zhou, Z.~Mo, Q.~Xiao, S.~Chen, and Y.~Yin, ``Privacy-preserving
  transportation traffic measurement in intelligent cyber-physical road
  systems,'' \emph{IEEE Transactions on Vehicular Technology}, vol.~65, no.~5,
  pp. 3749--3759, 2015.

\bibitem{ref-56}
B.~Hoh, M.~Gruteser, R.~Herring, J.~Ban, D.~Work, J.-C. Herrera, A.~M. Bayen,
  M.~Annavaram, and Q.~Jacobson, ``Virtual trip lines for distributed
  privacy-preserving traffic monitoring,'' in \emph{Proceedings of the 6th
  international conference on Mobile systems, applications, and services},
  2008, pp. 15--28.

\bibitem{ref-57}
K.~Xie, X.~Ning, X.~Wang, S.~He, Z.~Ning, X.~Liu, J.~Wen, and Z.~Qin, ``An
  efficient privacy-preserving compressive data gathering scheme in wsns,''
  \emph{Information Sciences}, vol. 390, pp. 82--94, 2017.

\bibitem{ref-10}
C.~{Dwork}, G.~N. {Rothblum}, and S.~{Vadhan}, ``Boosting and differential
  privacy,'' in \emph{2010 IEEE 51st Annual Symposium on Foundations of
  Computer Science}, Oct 2010, pp. 51--60.

\bibitem{ref-65}
F.~{Lyu}, N.~{Cheng}, H.~{Zhu}, H.~{Zhou}, W.~{Xu}, M.~{Li}, and X.~{Shen},
  ``Towards rear-end collision avoidance: Adaptive beaconing for connected
  vehicles,'' \emph{IEEE Transactions on Intelligent Transportation Systems},
  pp. 1--16, 2020.

\bibitem{ref-42}
Q.~Yang, Y.~Liu, T.~Chen, and Y.~Tong, ``Federated machine learning: Concept
  and applications,'' \emph{ACM Transactions on Intelligent Systems and
  Technology (TIST)}, vol.~10, no.~2, p.~12, 2019.

\bibitem{ref-38}
K.~Bonawitz, V.~Ivanov, B.~Kreuter, A.~Marcedone, H.~B. McMahan, S.~Patel,
  D.~Ramage, A.~Segal, and K.~Seth, ``Practical secure aggregation for
  federated learning on user-held data,'' \emph{arXiv preprint
  arXiv:1611.04482}, 2016.

\bibitem{ref-58}
G.~Xu, H.~Li, S.~Liu, K.~Yang, and X.~Lin, ``Verifynet: Secure and verifiable
  federated learning,'' \emph{IEEE Transactions on Information Forensics and
  Security}, vol.~15, pp. 911--926, 2019.

\bibitem{ref-59}
J.~Kang, Z.~Xiong, D.~Niyato, S.~Xie, and J.~Zhang, ``Incentive mechanism for
  reliable federated learning: A joint optimization approach to combining
  reputation and contract theory,'' \emph{IEEE Internet of Things Journal},
  vol.~6, no.~6, pp. 10\,700--10\,714, 2019.

\bibitem{ref-60}
J.~Ni, X.~Lin, and X.~S. Shen, ``Toward edge-assisted internet of things: from
  security and efficiency perspectives,'' \emph{IEEE Network}, vol.~33, no.~2,
  pp. 50--57, 2019.

\bibitem{ref-39}
T.~Nishio and R.~Yonetani, ``Client selection for federated learning with
  heterogeneous resources in mobile edge,'' in \emph{ICC 2019-2019 IEEE
  International Conference on Communications (ICC)}.\hskip 1em plus 0.5em minus
  0.4em\relax IEEE, 2019, pp. 1--7.

\bibitem{ref-40}
Y.~Chen, J.~Wang, C.~Yu, W.~Gao, and X.~Qin, ``Fedhealth: A federated transfer
  learning framework for wearable healthcare,'' \emph{arXiv preprint
  arXiv:1907.09173}, 2019.

\bibitem{ref-41}
S.~J. Pan and Q.~Yang, ``A survey on transfer learning,'' \emph{IEEE
  Transactions on knowledge and data engineering}, vol.~22, no.~10, pp.
  1345--1359, 2009.

\bibitem{ref-18}
T.~Li, A.~K. Sahu, A.~Talwalkar, and V.~Smith, ``Federated learning:
  Challenges, methods, and future directions,'' \emph{arXiv preprint
  arXiv:1908.07873}, 2019.

\bibitem{ref-12}
T.~Li, M.~Sanjabi, and V.~Smith, ``Fair resource allocation in federated
  learning,'' 2019.

\bibitem{ref-8}
Y.~Zhao, J.~Zhao, L.~Jiang, R.~Tan, and D.~Niyato, ``Mobile edge computing,
  blockchain and reputation-based crowdsourcing iot federated learning: A
  secure, decentralized and privacy-preserving system,'' \emph{arXiv preprint
  arXiv:1906.10893}, 2019.

\bibitem{ref-17}
Y.~Zhao, M.~Li, L.~Lai, N.~Suda, D.~Civin, and V.~Chandra, ``Federated learning
  with non-iid data,'' \emph{arXiv preprint arXiv:1806.00582}, 2018.

\bibitem{ref-50}
J.~Kang, Z.~Xiong, D.~Niyato, D.~Ye, D.~I. Kim, and J.~Zhao, ``Toward secure
  blockchain-enabled internet of vehicles: Optimizing consensus management
  using reputation and contract theory,'' \emph{IEEE Transactions on Vehicular
  Technology}, vol.~68, no.~3, pp. 2906--2920, 2019.

\bibitem{ref-14}
K.~Cho, B.~Van~Merri{\"e}nboer, C.~Gulcehre, D.~Bahdanau, F.~Bougares,
  H.~Schwenk, and Y.~Bengio, ``Learning phrase representations using rnn
  encoder-decoder for statistical machine translation,'' \emph{arXiv preprint
  arXiv:1406.1078}, 2014.

\bibitem{ref-19}
K.~Bonawitz, H.~Eichner, W.~Grieskamp, D.~Huba, A.~Ingerman, V.~Ivanov,
  C.~Kiddon, J.~Konečný, S.~Mazzocchi, H.~B. McMahan, T.~V. Overveldt,
  D.~Petrou, D.~Ramage, and J.~Roselander, ``Towards federated learning at
  scale: System design,'' 2019.

\bibitem{ref-21}
Wagstaff, Cardie, Claire, Rogers, Seth, and Stefan, ``Constrained k-means
  clustering with background knowledge,'' \emph{ICML-2001}, 2001.

\bibitem{ref-13}
C.~Chao, \emph{Freeway performance measurement system (pems)}, 2003.

\bibitem{ref-23}
T.~Ryffel, A.~Trask, M.~Dahl, B.~Wagner, J.~Mancuso, D.~Rueckert, and
  J.~Passerat-Palmbach, ``A generic framework for privacy preserving deep
  learning,'' \emph{arXiv preprint arXiv:1811.04017}, 2018.

\bibitem{pytorch}
A.~Paszke, S.~Gross, S.~Chintala, and G.~Chanan, ``Pytorch,'' \emph{Computer
  software. Vers. 0.3}, vol.~1, 2017.

\bibitem{ref-54}
M.~A. Mohandes, T.~O. Halawani, S.~Rehman, and A.~A. Hussain, ``Support vector
  machines for wind speed prediction,'' \emph{Renewable Energy}, vol.~29,
  no.~6, pp. 939--947, 2004.

\bibitem{ref-47}
Y.~Zhao, M.~Li, L.~Lai, N.~Suda, D.~Civin, and V.~Chandra, ``Federated learning
  with {Non-IID} data,'' 2018.

\bibitem{ref-52}
J.~Kang, Z.~Xiong, D.~Niyato, Y.~Zou, Y.~Zhang, and M.~Guizani, ``Reliable
  federated learning for mobile networks,'' \emph{arXiv preprint
  arXiv:1910.06837}, 2019.

\bibitem{ref-48}
T.~Li, A.~K. Sahu, A.~Talwalkar, and V.~Smith, ``Federated learning:
  Challenges, methods, and future directions,'' 2019.

\bibitem{ref-49}
T.~Li, A.~K. Sahu, M.~Zaheer, M.~Sanjabi, A.~Talwalkar, and V.~Smith,
  ``Federated optimization in heterogeneous networks,'' 2018.

\bibitem{ref-44}
T.~N. Kipf and M.~Welling, ``Semi-supervised classification with graph
  convolutional networks,'' \emph{arXiv preprint arXiv:1609.02907}, 2016.

\end{thebibliography}

%








\end{document}